%% file: motion.tex
\documentclass[10pt,journal,compsoc]{IEEEtran}
\ifCLASSOPTIONcompsoc
  \usepackage[nocompress]{cite}
\else
  \usepackage{cite}
\fi
\input{packages}
\input{macros}

\begin{document}
%
\title{Visual Dynamics: Stochastic Future Generation via Layered Cross Convolutional Networks}
%
%
%
%

\author{Tianfan~Xue*, 
        Jiajun~Wu*, 
        Katherine~L.~Bouman, 
        and~William~T.~Freeman
\IEEEcompsocitemizethanks{
\IEEEcompsocthanksitem T. Xue is with Google Inc., Mountain View, CA, 94043 USA.\protect\\E-mail: tianfan.xue@gmail.com.
\IEEEcompsocthanksitem J. Wu is with the Department of Electrical Engineering and Computer Science, Massachusetts Institute of Technology, Cambridge, MA, 02139 USA. E-mail: jiajunwu@mit.edu.
\IEEEcompsocthanksitem K. L. Bouman is with Harvard University, Cambridge, MA, 02138 USA. E-mail: klbouman@gmail.com.
\IEEEcompsocthanksitem W. T. Freeman is with the Department of Electrical Engineering and Computer Science, Massachusetts Institute of Technology, and Google Inc., Cambridge, MA, 02139 USA. E-mail: billf@mit.edu.}
\thanks{Manuscript received XXXX.}
\thanks{T. Xue and J. Wu contributed equally to this work.}}

%
%

\markboth{IEEE TRANSACTIONS ON PATTERN ANALYSIS AND MACHINE INTELLIGENCE, VOL. X, NO. X, MMMMMMM YYYY}%
{XUE \MakeLowercase{\textit{et al.}}: Visual Dynamics: Probabilistic Future Frame Synthesis via Cross Convolutional Networks}
%



\IEEEtitleabstractindextext{%
\input{text/abstract}

\begin{IEEEkeywords}
future prediction, frame synthesis, probabilistic modeling, convolutional networks, cross convolution
\end{IEEEkeywords}
}

\maketitle

\IEEEdisplaynontitleabstractindextext

%
\IEEEpeerreviewmaketitle

\input{text/intro}
\input{text/related_work}
\input{text/formulation}
\input{text/method}
\input{text/evaluation}
\input{text/analysis}

\input{text/application}
\input{text/conclusion}
\input{text/appendix}


%



\ifCLASSOPTIONcompsoc
  \section*{Acknowledgments}
\else
  \section*{Acknowledgment}
\fi

We thank Zhijian Liu and Yining Wang for helpful discussions and anonymous reviewers for constructive comments. This work was supported by NSF Robust Intelligence 1212849, NSF Big Data 1447476, ONR MURI 6923196, Adobe, Shell Research, and a hardware donation from Nvidia.

\ifCLASSOPTIONcaptionsoff
  \newpage
\fi



%



\bibliographystyle{IEEEtran}
\bibliography{motion}

%

\begin{IEEEbiography}[{\includegraphics[width=1in,height=1.25in,clip,keepaspectratio]{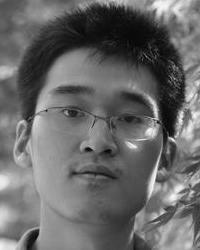}}]{Tianfan Xue} received his Ph.D. in EECS from MIT, working with William T. Freeman. Before that, he received his B.E. degree from Tsinghua Universtiy, and M.Phil. degree from The Chinese University of Hong Kong. His research interests include computer vision, image processing, and machine learning. Specifically, he is interested in motion estimation and image and video processing based on the motion information.
\end{IEEEbiography}

\begin{IEEEbiography}[{\includegraphics[width=1in,height=1.25in,clip,keepaspectratio]{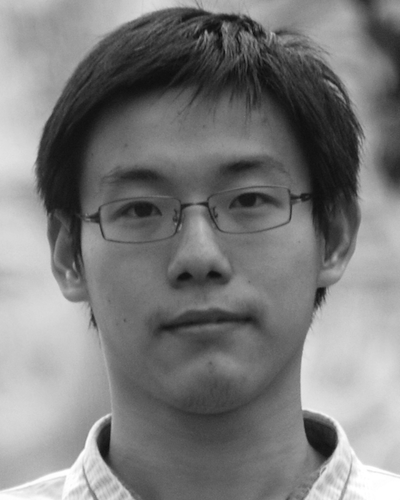}}]{Jiajun Wu} received the BEng degree in computer science from Tsinghua University, China, in 2014, and the SM degree in electrical engineering and computer science from MIT, in 2016. He is working toward the PhD degree in MIT, under the supervision of William T. Freeman and Joshua B. Tenenbaum. His research interests lie on the intersection of computer vision, machine learning, and computational cognitive science. He received the Facebook Fellowship, the Nvidia Fellowship, and the Adobe Fellowship. 
\end{IEEEbiography}

\begin{IEEEbiography}[{\includegraphics[width=1in,height=1.25in,clip,keepaspectratio]{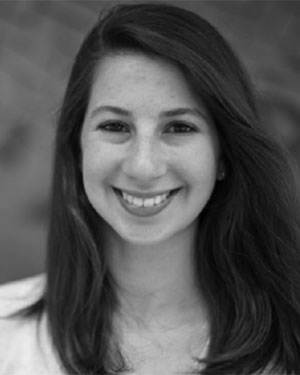}}]{Katherine L. Bouman} received the BSE degree in electrical engineering from the University of Michigan, Ann Arbor, Michigan, in 2011, and the SM and PhD degrees in electrical engineering and computer science from Massachusetts Institute of Technology (MIT), Cambridge, Massachusetts, in 2013 and 2017, under the supervision of William T. Freeman. Her master’s thesis on estimating material properties of fabric from video won the Ernst A. Guillemin Thesis Prize for outstanding S.M. thesis in electrical engineering with MIT. She received the US National Science Foundation Graduate Fellowship, the Irwin Mark Jacobs and Joan Klein Jacobs Presidential Fellowship, and is a Goldwater Scholar. Her research interests include computer vision, computational photography, and computational imaging.
\end{IEEEbiography}

\begin{IEEEbiography}[{\includegraphics[width=1in,height=1.25in,clip,keepaspectratio]{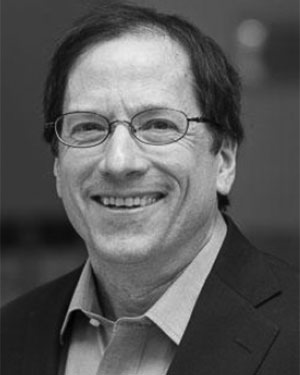}}]{William T. Freeman} is the Thomas and Gerd Perkins professor of electrical engineering and computer science with MIT, and a member of the Computer Science and Artificial Intelligence Laboratory (CSAIL) there. He was an associate department head from 2011-2014. His current research interests include machine learning applied to computer vision, Bayesian models of visual perception,and computational photography. He received outstanding paper awards at computer vision or machine learning conferences in 1997, 2006, 2009 and 2012, and test-of-time awards for papers from 1990 and 1995. Previous research topics include steerable filters and pyramids, orientation histograms, the generic viewpoint assumption, color constancy, computer vision for computer games, and belief propagation in networks with loops. He is active in the program or organizing committees of computer vision, graphics, and machine learning conferences. He was the program co-chair for ICCV 2005, and for CVPR 2013.
\end{IEEEbiography}







\end{document}

%% file: packages.tex
\usepackage{color,xcolor}
\usepackage{epsfig}
\usepackage{graphicx}

\usepackage[export]{adjustbox}
\usepackage{array}
\usepackage{booktabs}
\usepackage{colortbl}
\usepackage{float,wrapfig}
\usepackage{hhline}
\usepackage{multirow}
\usepackage{ragged2e}

\usepackage{amsmath,amsfonts,amsthm,amssymb}
\usepackage{bm}
\usepackage{nicefrac}
\usepackage{microtype}

\usepackage{changepage}
\usepackage{extramarks}
\usepackage{fancyhdr}
\usepackage{lastpage}
\usepackage{setspace}
\usepackage{soul}
\usepackage{xspace}

\usepackage[pagebackref=true,breaklinks=true,colorlinks]{hyperref}
\usepackage{url}

\usepackage{algorithm, algorithmic}
\usepackage{enumerate}
\usepackage{todonotes}

\usepackage{bbm}

%% file: macros.tex
\newcolumntype{L}[1]{>{\raggedright\let\newline\\\arraybackslash\hspace{0pt}}m{#1}}
\newcolumntype{C}[1]{>{\centering\let\newline\\\arraybackslash\hspace{0pt}}m{#1}}
\newcolumntype{R}[1]{>{\raggedleft\let\newline\\\arraybackslash\hspace{0pt}}m{#1}}
\newcommand{\specialcell}[2][c]{%
  \begin{tabular}[#1]{@{}l@{}}#2\end{tabular}}

\newcommand{\sect}[1]{Section~\ref{#1}}

\newcommand{\eqn}[1]{Eq.~(\ref{#1})}
\newcommand{\fig}[1]{Fig.~\ref{#1}}
\newcommand{\tbl}[1]{Table~\ref{#1}}

\newcommand{\ignorethis}[1]{}

\makeatletter
\DeclareRobustCommand\onedot{\futurelet\@let@token\@onedot}
\def\@onedot{\ifx\@let@token.\else.\null\fi\xspace}

\def\eg{\emph{e.g}\onedot} 
\def\ie{\emph{i.e}\onedot} 
\def\cf{\emph{c.f}\onedot}

\def\aka{a.k.a\onedot} \def\etal{\emph{et al}\onedot}
\makeatother

\definecolor{MyDarkBlue}{rgb}{0,0.08,1}
\definecolor{MyDarkGreen}{rgb}{0.02,0.6,0.02}
\definecolor{MyDarkRed}{rgb}{0.8,0.02,0.02}
\definecolor{MyDarkOrange}{rgb}{0.40,0.2,0.02}
\definecolor{MyPurple}{RGB}{111,0,255}
\definecolor{MyRed}{rgb}{1.0,0.0,0.0}
\definecolor{MyGold}{rgb}{0.75,0.6,0.12}
\definecolor{MyDarkgray}{rgb}{0.66, 0.66, 0.66}

\newcommand{\ignore}[1]{}

\def\cL{\mathcal{L}}
\def\i{^{(i)}}
\def\Netname{Probabilistic frame prediction\xspace}
\def\gendis{conditional data distribution\xspace}
\def\regdis{variational distribution\xspace}
\def\genmodel{generative model\xspace}
\def\regmodel{recognition model\xspace}
\def\zname{compact motion representation\xspace}
\def\ltwo{$\ell^2$}

\def\qphi{q_\phi(z|v\i,I\i)}
\def\pz{p_z(z)}
\def\pv{p(v^{(i)}|I\i)}
\def\ptheta{p_{\theta}(v\i|I\i,z)}
\def\pzvi{p(z|v\i,I\i)}

\newcommand{\myparagraph}[1]{\vspace{5pt}\noindent\textbf{#1}\hspace{10pt}}


%% file: text/abstract.tex
\begin{abstract}

We study the problem of synthesizing a number of likely future frames from a single input image. In contrast to traditional methods that have tackled this problem in a deterministic or non-parametric way, we propose to model future frames in a probabilistic manner. Our probabilistic model makes it possible for us to sample and synthesize many possible future frames from a single input image. To synthesize realistic movement of objects, we propose a novel network structure, namely a \emph{Cross Convolutional Network}; this network encodes image and motion information as feature maps and convolutional kernels, respectively. In experiments, our model performs well on synthetic data, such as 2D shapes and animated game sprites, and on real-world video frames. We present analyses of the learned network representations, showing it is implicitly learning a compact encoding of object appearance and motion. We also demonstrate a few of its applications, including visual analogy-making and video extrapolation.

\end{abstract}

%% file: text/intro.tex
\IEEEraisesectionheading{\section{Introduction}\label{sec:introduction}}

%
%
%
%

\IEEEPARstart{F}{rom} just a single snapshot, humans are often able to imagine multiple possible ways that the scene can change over time. For instance, due to the pose of the girl in \fig{fig:teaser}, most would predict that her arms are stationary but her leg is moving. However, the exact motion is often unpredictable due to an intrinsic ambiguity. Is the girl's leg moving up or down? How large is the movement?

In this work, we aim to depict the conditional distribution of future frames given a single observed image. We name the problem \emph{visual dynamics}, as it involves understanding how visual content relates to dynamic motions. We propose to tackle this problem using a probabilistic, content-aware motion prediction model that learns this distribution without using annotations. Sampling from this model allows us to visualize the many possible ways that an input image is likely to change over time.

The visual dynamics problem is in contrast to two traditional ways to model motion. The first is to assume object motion is deterministic, and to learn the direct mapping from an image's visual appearance to its motion~\cite{mathieu2015deep,Walker2014}. These methods are more likely to produce accurate results when there is little ambiguity in object motion (\eg, when long-range videos are available). For single image future prediction, however, the results are unlikely to align with reality. The second way to model motion is to derive its prior distribution, which is invariant to image content~\cite{fleet2000design,Weiss1998}. Understanding motion priors, like understanding image priors~\cite{Roth2005}, is a problem of fundamental scientific interest. However, as the motion we observe in real life is strongly correlated with visual content, methods ignoring visual signals do not work well on future frame prediction.

Modeling content-aware motion distributions is highly challenging mostly for two reasons. First, it involves modeling the correlation between motion and image content; to predict plausible future motion, the model must be able to relate visual content to semantic knowledge (\eg, object parts) in order to generate reasonable motions. Second, natural images lie on a high dimensional manifold that is difficult to describe precisely. Despite recent progresses on applying deep learning methods for image synthesis~\cite{radford2015unsupervised}, building a generative model for real images is far from being solved.

\input{figText/teaser}

We tackle the first problem using a novel convolutional neural network. During training, the network observes a set of consecutive image pairs from videos, and automatically infers the relationship between them without any supervision. During testing, the network then predicts the conditional distribution, $P(J|I)$, of future RGB images $J$ (\fig{fig:teaser}b) given an RGB input image $I$ that was not in the training set (\fig{fig:teaser}a). Using this distribution, the network is able to synthesize multiple different image samples corresponding to possible future frames for the input image (\fig{fig:teaser}c). 

We use a conditional variational autoencoder to model the complex conditional distribution of future frames~\cite{kingma2013auto,yan2015attribute2image}. This allows us to approximate a sample, $J$, from the distribution of future images by using a trainable function $J = f(I,z)$. The argument $z$ is a sample from a simple (\eg, Gaussian) distribution, which introduces randomness into the sampling of $J$.  This formulation makes the problem of learning the distribution more tractable than explicitly modeling the distribution.

To synthesize complex movement of objects, we proposed a novel layer-based synthesis network. The network splits an image into multiple segments and then uses a layered model to predict how each segment moves. This is a much easier task than modeling the motion of an entire image. Note that here we call each layer of an image as a segment to avoid confusion with convolutional layers. This layered prediction model synthesizes motion using a novel cross-convolutional layer. Unlike in standard convolutional layers, the values of the kernels are image-dependent, as different images may have different motions. Our model has several advantages. First, avoiding blurry outputs, the model does not directly synthesize the output image, but instead transforms pixels in the input frame based on sampled motion parameters. Second, the model only samples the movement of each layer, instead of all the pixels or a dense flow field. Because the motion of each layer lies on a lower-dimension manifold, its distribution is easier to model and the network can sample more diverse and realistic motions.

We test the proposed model on four datasets. Given an RGB input image, the algorithm can correctly model the distribution of possible future frames, and generate different samples that cover a variety of realistic motions. Our system significantly outperforms baselines in quantitative evaluation, and our results are in general preferred by humans in user studies.

We present analyses to reveal the knowledge captured by our model: the cross convolutional network is able to discover semantically meaningful parts that have coherent motion patterns in an unsupervised fashion; and the latent representation $z$ in the variational autoencoder is in essence a compact, interpretable encoding of object motion, as visualized in \sect{sec:analysis}. Our model has wide applications: we demonstrate that it can be applied to visual analogy-making and video extrapolation straightforwardly, with good qualitative and quantitative performance.

%% file: figText/teaser.tex
\begin{figure*}[t]
    \centering
    \includegraphics[width=\textwidth]{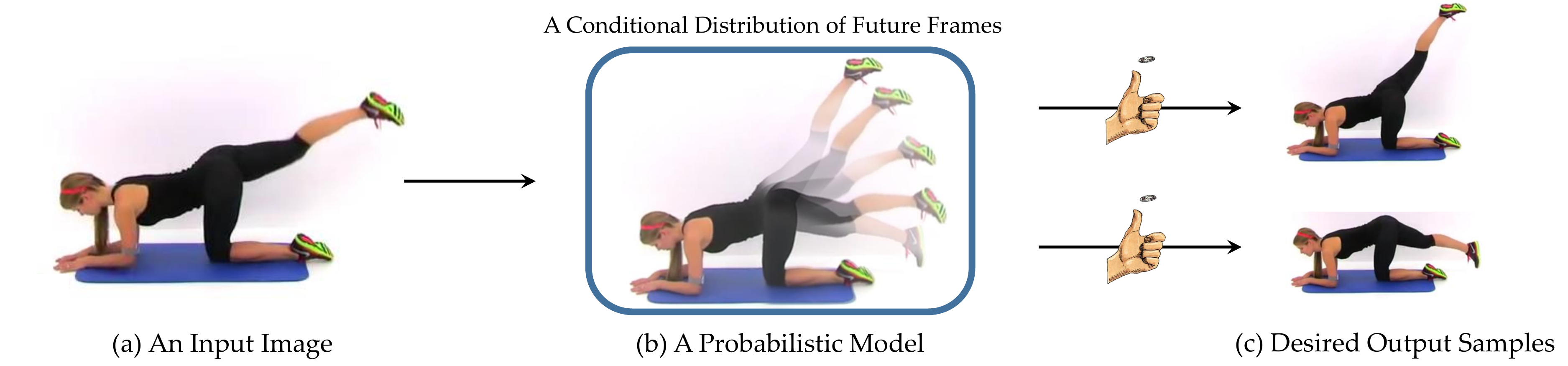}
    \caption{Predicting the movement of an object from a single snapshot is often ambiguous. For instance, is the girl's leg in (a) moving up or down? We propose a probabilistic, content-aware motion prediction model (b) that learns the conditional distribution of future frames, and produces a probable set of future frames (c). This schematic illustrates the idea behind our method, but does not show actual results produced by our model.}
    \label{fig:teaser}
\end{figure*}

%% file: text/related_work.tex
\section{Related Work}
\label{sec:related_work}

\myparagraph{Motion priors.}
Research studying the human visual system and motion priors provides evidence for low-level statistics of object motion. Pioneering work by Weiss and Adelson~\cite{Weiss1998} found that the human visual system prefers slow and smooth motion fields. More recent work by Lu and Yuille~\cite{Lu2006} found that humans make similar motion predictions as a Bayesian ideal observer. Roth and Black~\cite{Roth2005} analyzed the response of spatial filters applied to optical flow fields. Fleet~\etal\cite{fleet2000design} also found that a local motion field can be represented by a linear combination of a small number of bases. All these works focus on the distribution of a motion field itself without considering any image information. On the contrary, our context-aware model captures the relationship between an observed image and its motion field.

These prior works focused on modeling the distribution of an image's motion field using low-level statistics without any additional information. In real life, however, the distribution of motion fields is not independent of image content. For example, given an image of a car in front of a building, many would predict that the car is moving and the building is fixed. Thus, in our work, rather than modeling a motion prior as a context-free distribution, we propose to model the {\it conditional} motion distribution of future frames given an input image by incorporating a series of low- and high-level visual cues. 

\myparagraph{Motion or future prediction.} 
Given an observed image or a short video sequence, models have been proposed to predict a future motion field~\cite{liu2011sift,Pintea2014,xue2014refraction,Walker2015,WalkerDGH16}, future trajectories of objects~\cite{Walker2014,wu2015galileo,vda,uai,phys101}, or a future visual representation~\cite{vondrickanticipating}. However, unlike in our proposed model, most of these works use a deterministic prediction model~\cite{Pintea2014,vondrickanticipating}, which cannot model the uncertainty of the future.

Concurrently, Walker~\etal\cite{WalkerDGH16} identified the intrinsic ambiguity in deterministic prediction, and has proposed a probabilistic prediction framework. Our model is also probabilistic, but it directly predicts the pixel values rather than motion fields or image features.

\myparagraph{Parametric image synthesis.}
Early work in parametric image synthesis mostly focused on texture synthesis using hand-crafted features~\cite{portilla2000parametric}. More recently, works in image synthesis have begun to produce impressive results by training variants of neural network structures to produce novel images~\cite{gregor2015draw,xie2016synthesizing,xie2016deep3d,zhou2016view}. Generative adversarial networks~\cite{goodfellow2014generative,denton2015deep,radford2015unsupervised} and variational autoencoders~\cite{kingma2013auto,yan2015attribute2image} have been used to model and sample from natural image distributions. Our proposed algorithm is also based on the variational autoencoder, but unlike previous works, we also model the temporal consistency between frames.

\input{figText/pgm}

\myparagraph{Video synthesis.} 
Techniques that exploit the periodic structure of motion in videos have also been successful at generating novel frames from an input sequence. Sch{\"{o}}dl \etal proposed to shuffle frames from an existing video to generate a temporally consistent, looping image sequence~\cite{schodl2000video}. This idea was later used in video inpainting~\cite{wexler2004space}, and was extended to generate cinemagraphs~\cite{joshi2012cliplets} and seamlessly looping videos containing a variety of objects with different motion patterns~\cite{agarwala2005panoramic,liao2013automated}. While these techniques are able to generate high-resolution and realistic-looking videos, they are often applicable only to videos with periodic motions, and they require a reference video as input. In contrast, we build an image generation model that does not require a reference video during testing.

Recently, several neural network architectures have been proposed to synthesize a new frame from observed frames. They infer the future motion either from multiple previous frames~\cite{srivastava2015unsupervised,mathieu2015deep}, user-supplied action labels~\cite{oh2015action,finn2016unsupervised}, or directly model the joint distribution of all frames without conditioning on the input~\cite{vondrick2016generating}. In contrast to these approaches, our network takes a single frame as input and learns the conditional distribution of future frames without any supervision.

%% file: figText/pgm.tex
\begin{figure*}[t]
    \centering
    \includegraphics[width=\linewidth]{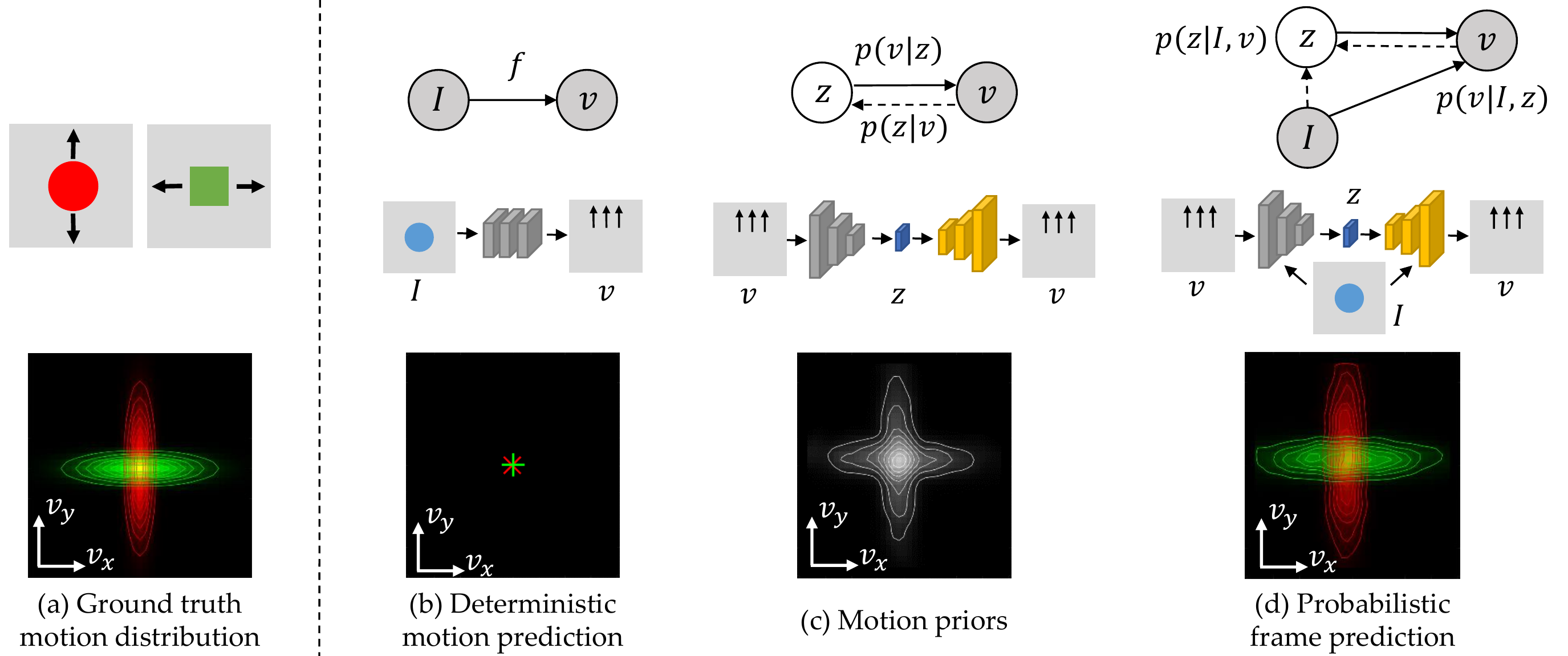}
    \caption{A toy example. Imagine a world composed of circles that move mostly vertically and squares that move mostly horizontally (a). We consider three different models (b-d) to learn the mapping from an image to a motion field. The first row shows graphical models, the second row shows corresponding network structures, and the third row shows estimated motion distributions. The deterministic motion prediction structure shown in (b) attempts to learn a one-to-one mapping from appearance to motion, but is unable to model multiple possible motions of an object, and tends to predict a mean motion for each object (the third row of (b)). The content-agnostic motion prior structure shown in (c) is able to capture a low-dimensional representation of motion, but is unable to leverage cues from image appearance for motion prediction. Therefore, it can only recovers the joint distribution of all objects (third row of (c)). The content-aware probabilistic motion predictor (d) brings together the advantages of models of (b) and (c) and uses appearance cues along with motion modeling to predict a motion field from a single input image. Therefore, the estimated motion distribution is very close to the ground truth (compare the last row and (a) and (d)).}
    \label{fig:graphical_model}
\end{figure*}

%% file: text/formulation.tex
\section{Formulation}
\label{sec:problem}

In this section, we first present a rigorous definition for the visual dynamics problem. Using a toy example, we then discuss three approaches to this problem, and show how the approach we take in our proposed model is more suitable for the task than the other two. We further present how our approach could be realized with a conditional variational autoencoder.

\subsection{Problem Definition}

In this section, we describe how to sample future frames from a current observation image. Here we focus on next frame synthesis; given an RGB image $I$ observed at time $t$, our goal is to model the conditional distribution of possible frames observed at time $t+1$.

Formally, let $\{(I^{(1)}, J^{(1)}), \dots, (I^{(n)}, J^{(n)})\}$ be the set of image pairs in the training set, where $I^{(i)}$ and $J^{(i)}$ are images observed at two consecutive time steps. Using this data, our task is to model the distribution $p_{\theta}(J|I)$ of all possible next frames $J$ for a new, previously unseen test image $I$, and then to sample new images from this distribution ($\theta$ is the set of model parameters). In practice, we choose not to directly predict the next frame, but instead to predict the difference image $v=J-I$ between the observed frame $I$ and the future frame $J$  (also known as the Eulerian motion). The task is then to learn the conditional distribution $p_{\theta}(v|I)$ from a set of training pairs  $\{(I^{(1)}, v^{(1)}), \dots, (I^{(n)}, v^{(n)})\}$.

\subsection{A Toy Example}
\label{sec:toy_example}

To understand how to design a model to best characterize the conditional distribution of object motion, consider a simple toy world that only consists of circles and squares. Circles mostly move vertically, while squares mostly move horizontally. As shown in \fig{fig:graphical_model}a, the ground truth distribution of a circle is $(v_x,v_y) \sim N((0,0),(0.2,1))$ and the distribution of a square is $N((0,0),(1,0.2))$, where $N(\vec{\mu},\vec{\sigma})$ is a Gaussian distribution with mean equals to $\vec{\mu}$ and diagonal variation equal to $\vec{\sigma}$. Using this toy model, we discuss how each of the three models shown in \fig{fig:graphical_model}b-d is able to infer the underlying motion.

\myparagraph{Approach I: Deterministic motion prediction.} 
In this structure, the model tries to find a deterministic relationship between the input image $I$ and object motion $v$ (\fig{fig:graphical_model}b). In our toy world, $I \in \{\mbox{circle},\mbox{square}\}$ is simply the binary label of each possible object and $v$ is a 2D motion vector\footnote{Although in practice we choose $v$ to be the RGB intensity difference between consecutive frames ($v = I-J$), for this toy example we define $v$ as the 2D motion vector.}.

In order to evaluate this model, we generate a toy training set which consists of 160,000 samples as follows. For each sample, we first randomly generate the object label $I$ with equal probabilities of being circles or squares, and then sample the 2D motion vector of the object based on its label. The model is trained by minimizing the reconstruction error $\sum_i ||v^{(i)} - f(I^{(i)})||$ on this toy training set. The two other models we will soon introduce are also trained on this toy dataset.\footnote{The last row of \fig{fig:graphical_model} shows actual predictions by our model trained on this dataset.}

One drawback of this deterministic model is that it cannot capture the multiple possible motions that a shape can have. Essentially, the model can only learn the average motion of each object, $I$. The third row of \fig{fig:graphical_model}b shows the estimated motion of both circles (the red cross) and squares (the green cross). Since the both circles and squares have zero-mean, symmetric motion distributions, this method predicts a nearly static motion field for each input image.

\myparagraph{Approach II: Motion priors.}
A simple way to model the multiple possible motions of future frames is to use a variational autoencoder~\cite{kingma2013auto}, as shown in \fig{fig:graphical_model}c. This model contains a latent representation, $z$ that encodes the intrinsic dimensionality of the motion fields. The network that learns this intrinsic representation $z$ consists of two parts: an encoder network $f$ that maps the motion field $v$ to an intrinsic representation $z$ (the gray network in \fig{fig:graphical_model}c, which corresponds to $p(z|v)$), and a decoder network $g$ that maps the intrinsic representation $z$ to the motion field $v$ (the yellow network, which corresponds to $p(v|z)$). During training, the network learns the latent representation $z$ by minimizing the reconstruction error on the training set $\sum_i ||v^{(i)} - g(f(v^{(i)}))||$. 

A shortcoming of this model is that the network does not see the input image when predicting the motion field. Therefore, it can only learn a joint distribution of both objects, as illustrated the third row of \fig{fig:graphical_model}c. Thus, during test time, the model is not be able to disambiguate between the specific motion distribution of circles and squares.

\myparagraph{Approach III: \Netname.}
In this work, we combine the deterministic motion prediction structure (approach I) with a motion prior (approach II), to model the uncertainty in a motion field and the correlation between motion and image content. We extend the decoder in (2) to take two inputs, the intrinsic motion representation $z$ and an image $I$ (see the yellow network in \fig{fig:graphical_model}d, which corresponds to $p(v|I,z)$). Therefore, instead of solely being able to model a joint distribution of motion $v$, it is now able to learn a conditional distribution of motion given the input image $I$.

In this toy example, since squares and circles move primarily in one (although different) direction, the intrinsic motion representation $z$ only records the magnitude of motion along their major and minor directions. Combining the intrinsic motion representation with the direction of motion inferred from the image content, the model can correctly model the distribution of motion. \fig{fig:graphical_model}d shows that the inferred motion distribution of each object is quite similar to the ground truth distribution.

\input{figText/pipeline}

\subsection{Conditional Variational Autoencoder}
\label{sec:vae}

In this section, we will formally derive the training objective of our model, following the similar derivations~\cite{kingma2013auto,kingma2014semi,yan2015attribute2image}. Consider the following generative process that samples a future frame conditioned on an observed image, $I$. First, the algorithm samples the hidden variable $z$ from a prior distribution $p_z(z)$; in this work, we assume $p_z(z)$ is a multivariate Gaussian distribution where each dimension is {i.i.d.} with zero-mean and unit-variance. Then, given a value of $z$, the algorithm samples the intensity difference image $v$ from the conditional distribution $p_\theta(v|I,z)$. The final image, $J = I + v$, is then returned as output.

\myparagraph{A variational upper-bound.}
In the training stage, the algorithm attempts to maximize the log-likelihood of the conditional marginal distribution $\sum_i \log p(v^{(i)}|I\i)$. Assuming $I$ and $z$ are independent, the marginal distribution is expanded as $\sum_i \log \int_z p(v^{(i)}|I\i,z) p_z(z) dz$. Directly maximizing this marginal distribution is hard, thus we instead maximize its variational upper-bound~\cite{kingma2013auto}. Each term in the marginal distribution is upper-bounded by
\begin{align}
\cL(\theta,\phi,v\i|I\i) \approx & - D_{\text{KL}}(q_\phi(z|v\i,I\i) || p_z(z)) \nonumber \\
& + \frac{1}{L} \sum_{l=1}^{L} \left[\log p_\theta(v\i|z^{(i,l)},I\i)\right], 
\label{eq:sample_approx}
\end{align}
where $D_{\text{KL}}$ is the KL-divergence, $q_\phi(z|v\i,I\i)$ is the variational distribution that approximates the posterior $p(z|v\i,I\i)$, and $z^{(i,l)}$ are samples from the \regdis. Recall that $z$ and $I\i$ are independent, so that $p_z(z|I\i) = p_z(z)$. Please see \sect{sec:appendix} for detailed derivation of \eqn{eq:sample_approx}. For simplicity, we refer to the \gendis, $p_\theta(\cdot)$, as the {\it \genmodel}, and the \regdis, $q_\phi(\cdot)$, as the {\it\regmodel}. 

In practice, we always choose $L=1$. Therefore, the upper bound of the KL-divergence can be simplified as
\begin{equation}
- D_{\text{KL}}(q_\phi(z|v\i,I\i) || p_z(z)) + \log p_\theta(v\i|z^i,I\i).
\label{eq:sample_approx_sim}
\end{equation}

If the assumption that $I$ and $z$ are independent does not hold, we can convert a latent variable that depends on $I$ to one that does not, without affecting the expressiveness of our generative model, as suggested by Sohn~\etal~\cite{kingma2014semi,sohn2015learning}.

\myparagraph{Distribution reparametrization.}
We assume Gaussian distributions for both the \genmodel and \regmodel\footnote{A complex distribution can be approximated as a function of a simple distribution, such as a Gaussian. This is referred to as the reparameterization trick~\cite{kingma2013auto}.}, where the mean and variance of the distributions are functions specified by neural networks, that is\footnote{Here the bold $\mathbf{I}$ denotes an identity matrix, whereas the normal-font $I$ denotes the observed image.} 
\begin{align}
p_{\theta}(v\i|z^{(i,l)},I\i) &= \mathcal{N}(v\i; f_{\mbox{\tiny{mean}}}(z^{(i,l)},I\i), \sigma^2 \mathbf{I}), \label{eq:repara_p}\\
q_{\phi}(z^{(i,l)}|v\i,I\i) &= \mathcal{N}(z^{(i,l)}; g_{\mbox{\tiny{mean}}}(v\i,I\i), g_{\mbox{\tiny{var}}}(v\i,I\i)), \label{eq:repara_q}
\end{align}
where $\mathcal{N}(\;\cdot\;;a,b)$ is a Gaussian distribution with mean $a$ and variance $b$. $f_{\mbox{\tiny{mean}}}$ is a function that predicts the mean of the \genmodel, defined by the generative network (the yellow network in \fig{fig:graphical_model}d). $g_{\mbox{\tiny{mean}}}$ and $g_{\mbox{\tiny{var}}}$ are functions that predict the mean and variance of the \regmodel, respectively, defined by the recognition network (the gray network in \fig{fig:graphical_model}d). Here we assume that all dimensions of the \genmodel have the same variance $\sigma^2$, where $\sigma$ is a hand-tuned hyperparameter. 

\myparagraph{The objective function.} 
Plugging \eqn{eq:repara_p} and \eqn{eq:repara_q} in to \eqn{eq:sample_approx}, for $L=1$ (only use one sample for each training iteration) we obtain the objective function that we minimize for each sample:
\begin{align}
D_\text{KL}(q_\phi(z|v\i,I\i) || p_z(z)) + \lambda \|v\i - f_{\mbox{\tiny{mean}}}(z^{(i)},I\i)\|,
\label{eq:final_bound}
\end{align}
where $z^{(i)}$ is sampled from the distribution defined by \eqn{eq:repara_q}, and $\lambda$ is a constant. During training, we use stochastic gradient descent to minimize the variational lower bound defined in \eqn{eq:final_bound}. 

We will describe in \sect{sec:method} the neural networks that define the generative function $f_{\mbox{\tiny{mean}}}$ and recognition function $g_{\mbox{\tiny{mean}}}$ and $g_{\mbox{\tiny{var}}}$.

%% file: figText/pipeline.tex
\begin{figure*}[t]
    \centering
    \includegraphics[width=\linewidth]{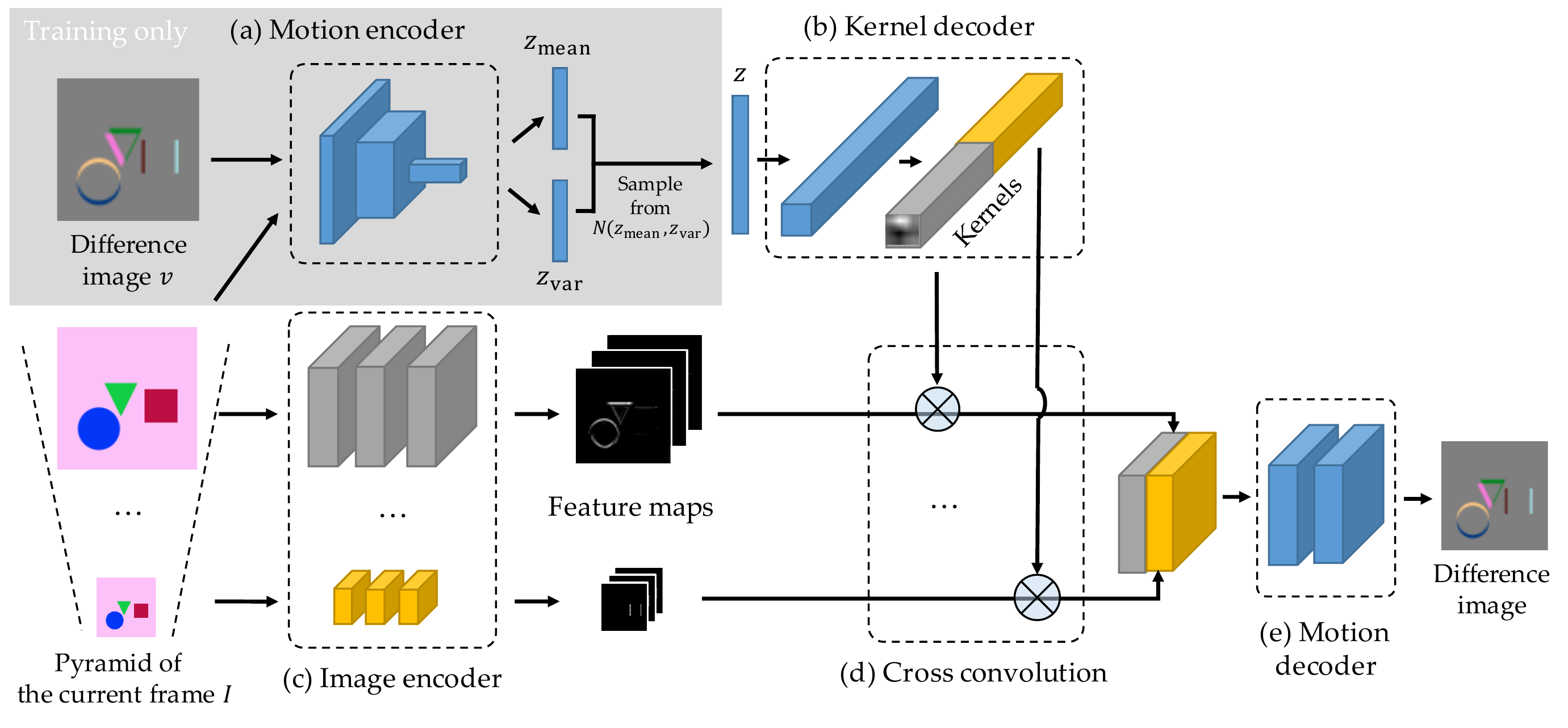}
    \caption{Our network consists of five components: (a) a motion encoder, (b) a kernel decoder, (c) an image encoder, (d) a cross convolution layer, and (e) a motion decoder. Our image encoder takes images at four scales as input. For simplicity, we only show two scales in this figure. See \sect{sec:method} for details of our model. The motion encoder (the grayed region) is only used in training. At testing time, the motion vector $z$ is sampled from its empirical distribution (see text for more details).}
    \label{fig:pipeline}
\end{figure*}

%% file: text/method.tex
\section{Method}
\label{sec:method}

We present an end-to-end trainable neural network structure, defining the generative function $f_{\mbox{\tiny{mean}}}$ and recognition functions $g_{\mbox{\tiny{mean}}}$, and $g_{\mbox{\tiny{var}}}$. Once trained, these functions can be used in conjunction with an input image to sample future frames. We first describe our newly proposed cross convolutional layer, which naturally characterizes a layered motion representation~\cite{wang1993layered}. We then explain our network structure and demonstrate how we integrate the cross convolutional layer into the network for future frame synthesis. 

\subsection{Layered Motion Representations and Cross Convolutional Networks}
\label{sec:crossconv}

Motion can often be decomposed in a layer-wise manner~\cite{wang1993layered}. Intuitively, different semantic segments in an image should have different distributions over all possible motions; for example, a building is often static, but a car moves.

To model layered motion, we propose a novel cross convolutional network (\fig{fig:pipeline}). The network first decomposes an input image pyramid into multiple feature maps (segments) through an image encoder (\fig{fig:pipeline}c). It then convolves these maps with different kernels (\fig{fig:pipeline}d), and uses the outputs to synthesize a difference image (\fig{fig:pipeline}e). This network structure naturally fits a layered motion representation, as each feature map characterizes an image \emph{segment} and the corresponding kernel characterizes the motion of that segment. In other words, we model motions as convolutional kernels, which are applied to feature maps of images at multiple scales. 

Unlike a traditional convolutional network, these kernels should not be identical for all inputs, as different images should be associated with different motions (kernels). We therefore propose a cross convolutional layer to tackle this problem. The cross convolutional layer does not learn the weights of the kernels itself. Instead, it takes both kernel weights and image segments as input and performs convolution during a forward pass; for back propagation, it computes the gradients with respect to both convolutional kernels and image segments.
 
The characteristics of a cross convolutional layer naturally fit the layered motion representation, as we can think of each feature map as an image segment, and the corresponding kernel characterizes the layer's motion. In other words, we model motions as convolutional kernels, which are applied to image segments (layers) at multiple scales. Concurrent papers~\cite{finn2016unsupervised,de2016dynamic} have also explored similar ideas. While they applied the learned kernels on input images, we jointly learn image segments and kernels without direct supervision.

\subsection{Network Structure}

As shown in \fig{fig:pipeline}, our network consists of five components: (a) a motion encoder, which is a variational autoencoder that learns the compact representation, $z$, of possible motions; (b) a kernel decoder, which learns the motion kernels from the compact motion representation $z$; (c) an image encoder, which consists of convolutional layers extracting segments from the input image $I$; (d) a cross convolutional layer, which takes the output of the image encoder and the kernel decoder, and convolves the image segments with motion kernels; and (e) a motion decoder, which regresses the difference image from the combined feature maps. We now introduce each part in detail.

During training, our motion encoder (\fig{fig:pipeline}a) takes the current frame and a difference image as input, both at resolution 128$\times$128. The network then applies six 5$\times$5 convolutional and batch normalization layers (number of channels are $\{$96, 96, 128, 128, 256, 256$\}$) to the concatenated images, with some pooling layers in between. The output has a size of 256$\times$5$\times$5. The kernel encoder then reshapes the output to a vector, and splits it into a 3,200-dimension mean vectors $z_\text{mean}$ and a 3,200-dimension variance vector $z_\text{var}$, from which the network samples the latent motion representation $z \sim N(z_\text{mean}, z_\text{var})$. The motion encoder takes the current frame as input, in addition to the motion image, so that it can learn to model the \emph{conditional} variational distribution ($q_\theta(\cdot)$ in \eqn{eq:final_bound}).

Next, the kernel decoder (\fig{fig:pipeline}b) sends the 3,200 $=$ 128$\times$5$\times$5 tensor into two additional convolutional layers, each with 128 channels and a kernel size of 5. They are then split into four sets, each with 32 kernels of size 5$\times$5. 

Our image encoder (\fig{fig:pipeline}c) operates on four different scaled versions of the input image $I$ (256$\times$256, 128$\times$128, 64$\times$64, and 32$\times$32)\footnote{For the input image of size $128 \times 128$, we used five different scales instead. In that case, the size of motion vector is 4,000 ($=5 \times 5 \times 32 \times 5$).}. At each scale, there are four sets of 5$\times$5 convolutional and batch normalization layers (number of channels are $\{$64, 64, 64, 32$\}$), two of which are followed by a 2$\times$2 max pooling layer. Therefore, the output size of the four channels are 32$\times$64$\times$64, 32$\times$32$\times$32, 32$\times$16$\times$16, and 32$\times$8$\times$8, respectively. This multi-scale convolutional network allows us to model both global and local structures in the image, which may have different motions. 

The core of our network is a cross convolutional layer (\fig{fig:pipeline}d), which, as discussed in \sect{sec:crossconv}, applies the kernels learned by the kernel decoder to the feature maps (layers) learned by the image encoder. The cross convolutional layer has the same  output size as the image encoder. 

Our motion decoder (\fig{fig:pipeline}e) starts with an up-sampling layer at each scale, making the output of all scales of the cross convolutional layer have a resolution of 64$\times$64. This is then followed by one 9$\times$9 and two 1$\times$1 convolutional and batch normalization layers, with $\{$128, 128, 3$\}$ channels. These final feature maps (layers) are then used to regress the output difference image. 

\myparagraph{Training and testing details.} 
During training, the image encoder takes a single frame $I^{(i)}$ as input, and the motion encoder takes both input frame $I^{(i)}$ and the difference image $v^{(i)}=J^{(i)}-I^{(i)}$ as input, where $J\i$ is the next frame. The network aims to regress the difference image that minimizes the objective function~\eqn{eq:final_bound}. 

During testing, the image encoder still sees a single image $I$; however, instead of using a motion encoder, we directly sample motion vectors $z^{(j)}$ from the prior distribution $p_z(z)$ (therefore, the gray part in \fig{fig:pipeline} is not used in testing). In practice, we use an empirical distribution of $z$ over all training samples as an approximation to the prior, \aka the variational distribution $q_{\phi}(z)$ in the literature, as Doersch~\etal~\cite{doersch2016tutorial} show that this sampling strategy works better than sampling from the prior distribution $p_z(z)$. Our sampling of $z$ is independent of the input image $I$, satisfying the independence assumption discussed in \sect{sec:vae}. The network then synthesizes possible difference images $v^{(j)}$ by taking the sampled latent representation $z^{(j)}$ and an RGB image $I$ as input. We then generate a set of future frames $\{J^{(j)}\}$ from these difference images: $J^{(j)}=I+v^{(j)}$.

%% file: text/evaluation.tex
\section{Evaluations}
\label{sec:eva}

\input{figText/shape}

\input{figText/shape_analysis}

We now present a series of experiments to evaluate our method. We start with a dataset of 2D shapes, which serves to benchmark our model on objects with simple, yet nontrivial, motion distributions. Following Reed~\etal~\cite{reed2015deep}, we then test our method on a dataset of video game sprites with diverse motions. In addition to these synthetic datasets, we further evaluate our framework on real-world video datasets. Again, note that our model uses consecutive frames for training, requiring no supervision. Visualizations of our experimental results are also available on our project page\footnote{Our project page: \url{http://visualdynamics.csail.mit.edu}}.

\input{figText/game}
\subsection{Movement of 2D Shapes}
\label{sec:exp_shapes}

We first evaluate our method using a dataset of synthetic 2D shapes. This dataset serves to benchmark our model on objects with simple, yet nontrivial, motion distributions. It contains three types of objects: circles, squares, and triangles. Circles always move vertically, squares horizontally, and triangles diagonally. The motion of circles and squares are independent, while the motion of circles and triangles are correlated (when the triangle moves up, the circle moves down). The shapes can be heavily occluded, and their sizes, positions, and colors are chosen randomly. There are 20,000 pairs for training, and 500 for testing.

\fig{fig:result_shape} shows the results. \fig{fig:result_shape}a and \fig{fig:result_shape}b show a sample of consecutive frames in the dataset, and \fig{fig:result_shape}c shows the reconstruction of the second frame after encoding and decoding with the ground truth image pairs. \fig{fig:result_shape}d and \fig{fig:result_shape}e show samples of the second frame; in these results the network only takes the first image as input, and the \zname, $z$, is randomly sampled. Note that the network is able to capture the distinctive motion pattern for each shape, including the strong correlation of triangle and circle motion. 

To quantitatively evaluate our algorithm, we compare the displacement distributions of circles, squares, and triangles in the sampled images with their ground truth distributions. We sample 50,000 images and use the optical flow package by Liu~\etal~\cite{liu2009beyond} to calculate the mean movement of each object. We plot them in \fig{fig:result_dist} as well as the isolines using the \emph{contour} function in MATLAB. 

We also compute their KL-divergence. Here, we divide the region $[-5, 5] \times [-5, 5]$ into $41 \times 41$ bins and approximate the predicted distribution with the 2D histogram. We compare our algorithm with a simple baseline that copies the optical flow field of the closest image pairs from the training set (`Flow' in \fig{fig:result_dist}); for each test image, we find its 10-nearest neighbors in the training set  (the retrieval is based on \ltwo-distance between query image and images in the training dataset), and randomly transfer one of the corresponding optical flow fields. To illustrate the advantage of using a variational autoencoder (VAE) over a standard autoencoder, we also modify our network by removing the KL-divergence loss and sampling layer (`AE' in \fig{fig:result_dist}). \fig{fig:result_dist} shows our predicted distribution is very close to the ground-truth distribution. It also shows that a VAE helps to capture the true distribution of future frames.

\input{figText/cardio}
\subsection{Movement of Video Game Sprites}

We evaluate our framework on a video game sprites dataset\footnote{Liberated pixel cup: \url{http://lpc.opengameart.org}}, also used by Reed~\etal\cite{reed2015deep}. The dataset consists of 672 unique characters; for each character, there are 5 animations (spellcast, thrust, walk, slash, shoot) from $4$ different viewpoints. The length of each animation ranges from 6 to 13 frames. We collect 102,364 pairs of neighboring frames for training, and 3,140 pairs for testing. The same character does not appear in both the training and test sets. Sampled future frames are shown in \fig{fig:result_game}. From a single frame, our method captures various possible motions that are consistent with those in the training set.

As a quantitative evaluation on the success rate of our image synthesis algorithm, we conduct behavioral experiments on Amazon Mechanical Turk. We randomly select 200 images, sample a possible next frame using our algorithm, and show them to multiple human subjects as an animation side by side with the ground truth animation. We then ask the subject to choose which animation is real (not synthesized). An ideal algorithm should achieve a success rate of 50\%. In our experiments, we present the animation in both the original resolution (64$\times$64) and a lower resolution (32$\times$32). We only evaluate on subjects that have a past approval rate of $>$95\% and have also passed our qualification tests. \fig{fig:result_game} shows that our algorithm significantly outperforms a baseline that warps the input by transferring one of closest flow fields from the training set. Our results are labeled as \emph{real} by humans 41.2\% of the time at 32$\times$32, and 35.7\% of the time at 64$\times$64. Subjects are less easily fooled by 64$\times$64 images, as it is harder to hallucinate realistic details in high-resolution images.

\input{figText/mask_kernel}
\input{figText/action}

\subsection{Movement in Real Videos Captured in the Wild}

To demonstrate that our algorithm can also handle real videos, we collect 20 workout videos from YouTube, each about 30 to 60 minutes long. We first apply motion stabilization to the training data as a pre-processing step to remove camera motion. We then extract 56,838 pairs of frames for training and 6,243 pairs for testing. The training and testing pairs come from different video sequences. 

\fig{fig:result_cardio} shows that our framework works well in predicting the movement of the legs and torso. Specifically, the algorithm is able to synthesize reasonable motions of the human in various poses. The Mechanical Turk behavioral experiments also show that the synthesized frames are visually realistic. In particular, our synthesized images are labeled as real by humans 36.7\% of the time at a resolution of 32$\times$32, and 31.3\% of the time at 64$\times$64.

Given an input frame, how often can our algorithm sample realistic motion? We have run an additional human study to evaluate this: for each of 100 randomly selected test images, we sample 100 future frames at 128$\times$128, and ask AMT subjects whether the two consecutive frames, one real and one synthesized, characterize realistic human motion. The success rate is 44.3\% (averaged over the 100 test images), with a standard deviation of 4.3\%.

Synthesizing images with realistic human motion is challenging. Traditional methods often require a high quality 3D model of the human body for real-time synthesis and rendering~\cite{thies2016face2face}. Although our algorithm does not require a 3D model, it can still simulate how a human moves between frames. Note that synthesized large motions in our results often happen around locations corresponding to skeletal joints, implying that our model has an implicit understanding of the structure and correlation between body parts.

\input{figText/zstats_fig}

\input{figText/zstats_tbl}

At last, we test our algorithm on realistic videos in the wild. We use the PennAction dataset~\cite{zhang2013actemes}, which contains 981 sequences of diverse human motion with complex backgrounds. We extract 7,705 pairs of frames for training and 987 pairs for evaluation. To remove trivial panning motions, we detect the human in each frame~\cite{huang2017expecting} and crop each frame to center the subject.

The results are shown in \fig{fig:result_action}. While in-the-wild videos are more challenging, our model synthesizes multiple realistic future frames from a single image. Especially, when synthesizing future frames, our model keeps the original background intact, suggesting the learned motion representation separates the region of interest from the background. In the future, we aim to develop prediction models that are able to better handle more complex visual scenes.

%% file: figText/shape.tex
\begin{figure*}[t]
    \centering
    \footnotesize
    \includegraphics[width=\linewidth]{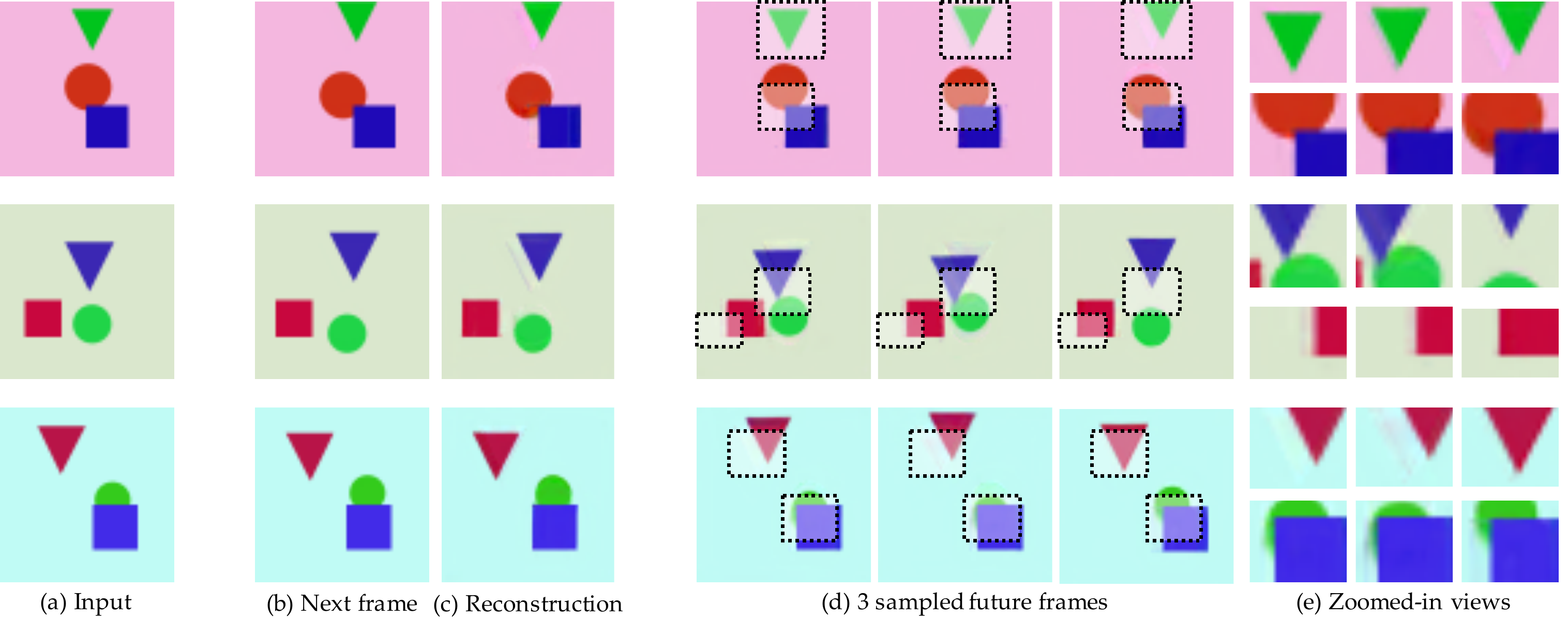}    
 
    \caption{Results on the \emph{Shapes} dataset containing circles, squares, and triangles. For a given frame (a) our goal is to predict probable motion. In (b) we show the ground truth future frame. Notice how squares move horizontally, circles vertically, triangles diagonally, and the triangle's motion is correlated with the circle's. Our model is able to reconstruct the motion (c) after encoding and decoding with the ground truth image pairs. By sampling from the latent representation, we can also synthesize additional novel future frames with probable motion (d). In (e), we show zoomed-in regions for these samples. Note the  significant variation among the sampled frames.
    }  
    \label{fig:result_shape}
\end{figure*}

%% file: figText/shape_analysis.tex
\begin{figure*}[t]
    \centering
    \begin{minipage}{0.5\linewidth}
        \includegraphics[width=\linewidth]{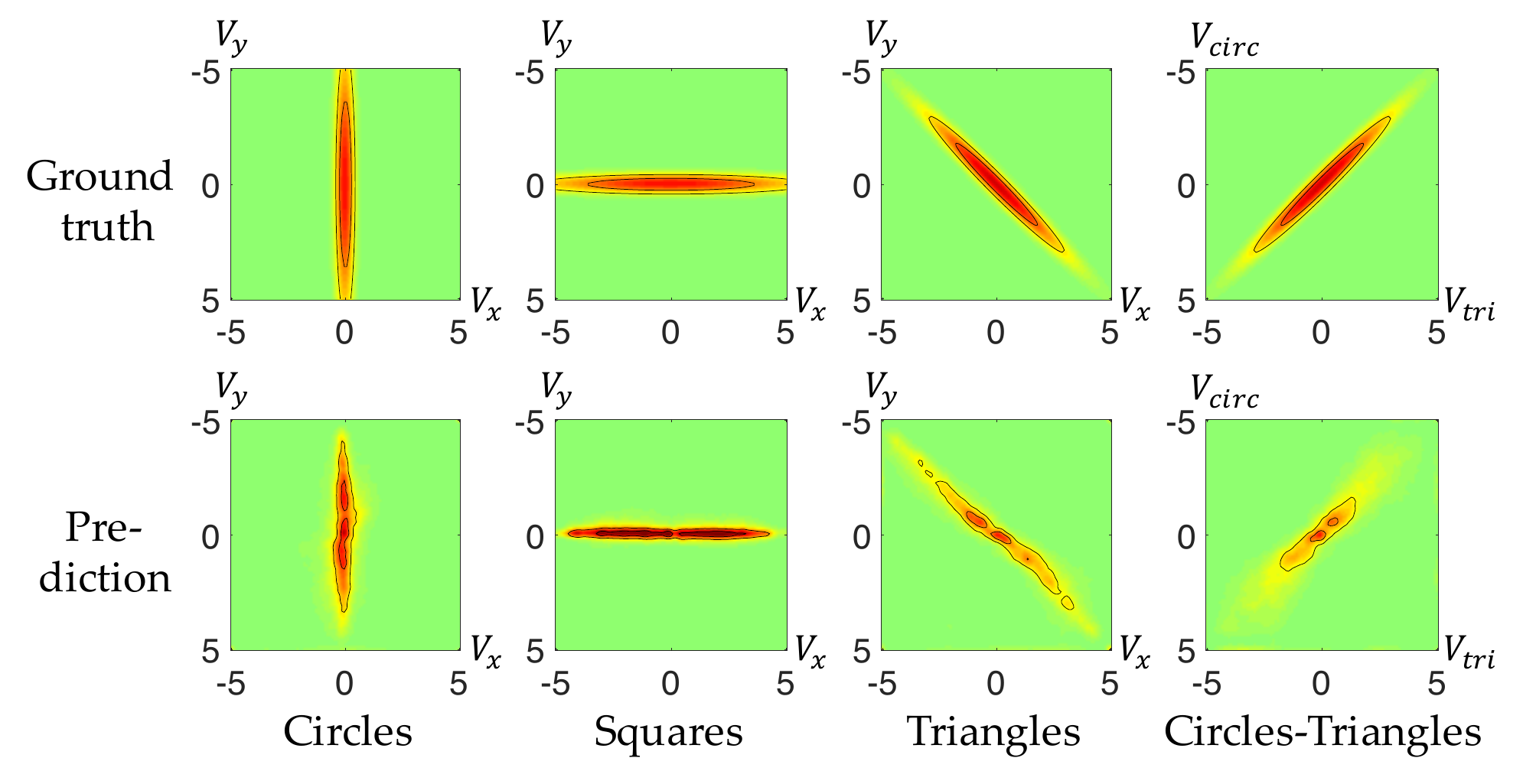} 
    \end{minipage}
    \begin{minipage}{0.49\linewidth}
        
        \renewcommand{\arraystretch}{1.3} 
        \setlength{\tabcolsep}{5pt}
        \begin{tabular}{lcccc}
            \toprule
            \multirow{2}{*}{Method} & \multicolumn{4}{c}{Shapes}\\ 
            \cmidrule{2-5}
             & Circles & Squares & Triangles & Circles-Triangles \\ 
            \midrule
            Flow & 6.77 & 7.07 & 6.07 & 8.42 \\
            AE & 8.76 & 12.37 & 10.36 & 10.58 \\
            Ours & \textbf{1.70} & \textbf{2.48} & \textbf{1.14} & \textbf{2.46} \\
            \bottomrule
        \end{tabular}
        
        \ \\
        \justifying\noindent
        \textit{\small KL divergence, $D_\text{KL}(p_{\text{gt}} \mid\mid p_{\text{pred}})$, between predicted and ground truth distributions}
    \end{minipage}
        
    \caption{The motion of the sampled data is consistent with the ground truth motion distributions. Left: for each object, comparison between its ground-truth motion distribution and the distribution predicted by our method. It shows the network learns to move circles vertically, squares, horizontally, and the motion of circles and triangles is correlated. Right: KL divergence between ground-truth distributions and distributions predicted by three different algorithms. Our network scores much better than a simple nearest-neighbor motion transfer algorithm. }
    \label{fig:result_dist}
\end{figure*}

%% file: figText/game.tex
\begin{figure*}[t]
    \centering
    \begin{minipage}{0.71\linewidth}
        \includegraphics[width=\linewidth]{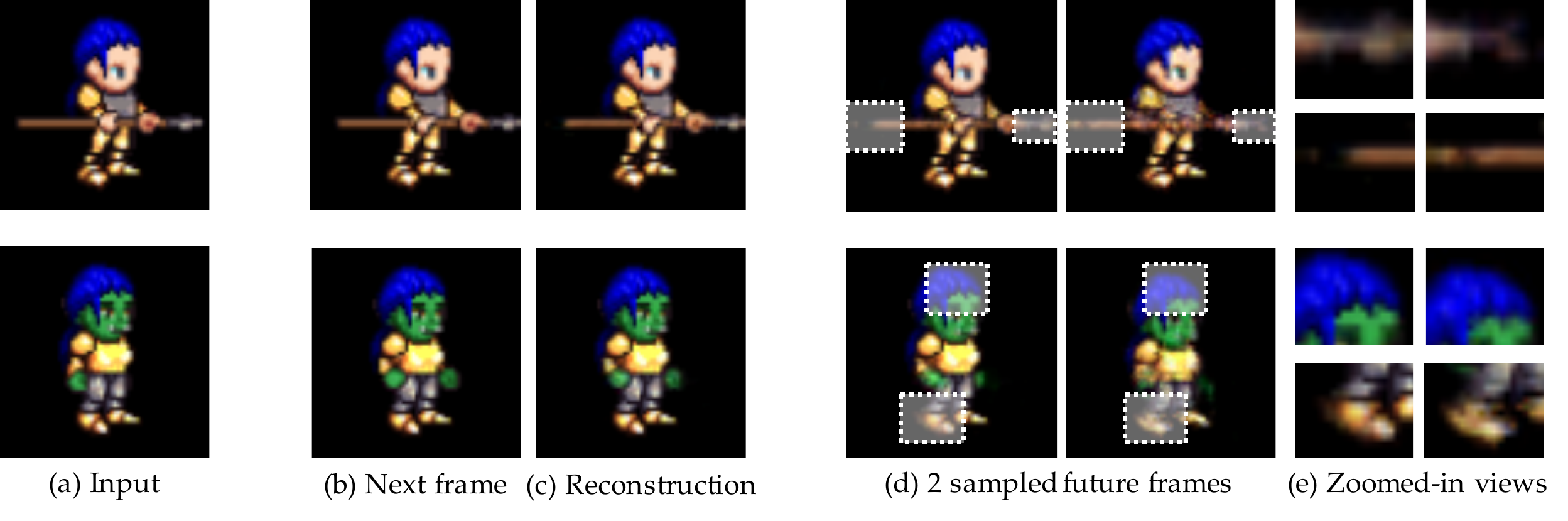} 
    \end{minipage}
    \hfill
    \begin{minipage}{0.27\linewidth}
        \centering
        \renewcommand{\arraystretch}{1.3} 
        \begin{tabular}{ccc}
            \toprule
            \multirow{2}{*}{Method} & \multicolumn{2}{c}{Resolution} \\
            \cmidrule{2-3}
            & 32$\times$32 & 64$\times$64 \\
            \midrule
            Flow & 29.7 & 21.0 \\
            Ours & \textbf{41.2} & \textbf{35.7} \\
            \bottomrule
        \end{tabular}
        
        \ \\
        \justifying\noindent
        \textit{\small Percentages (\%) of results labeled as \emph{real} by human subjects}
    \end{minipage}

    \caption{Left: Results on the \emph{Sprites} dataset, where we show input images (a), ground truth next frames (b), our reconstruction (c), two sampled future frames (d), and corresponding zoomed-in views (e). Right: Percentages (\%) of synthesized results that were labeled as real by human subjects in two-way forced choices on Amazon Mechanical Turk, at resolution 32$\times$32 and 64$\times$64. A perfect algorithm would achieve a percentage around 50\%.}
    \label{fig:result_game}
\end{figure*}

%% file: figText/cardio.tex
\begin{figure*}[t]
    \centering
        \begin{minipage}{0.70\linewidth}
        \includegraphics[width=\linewidth]{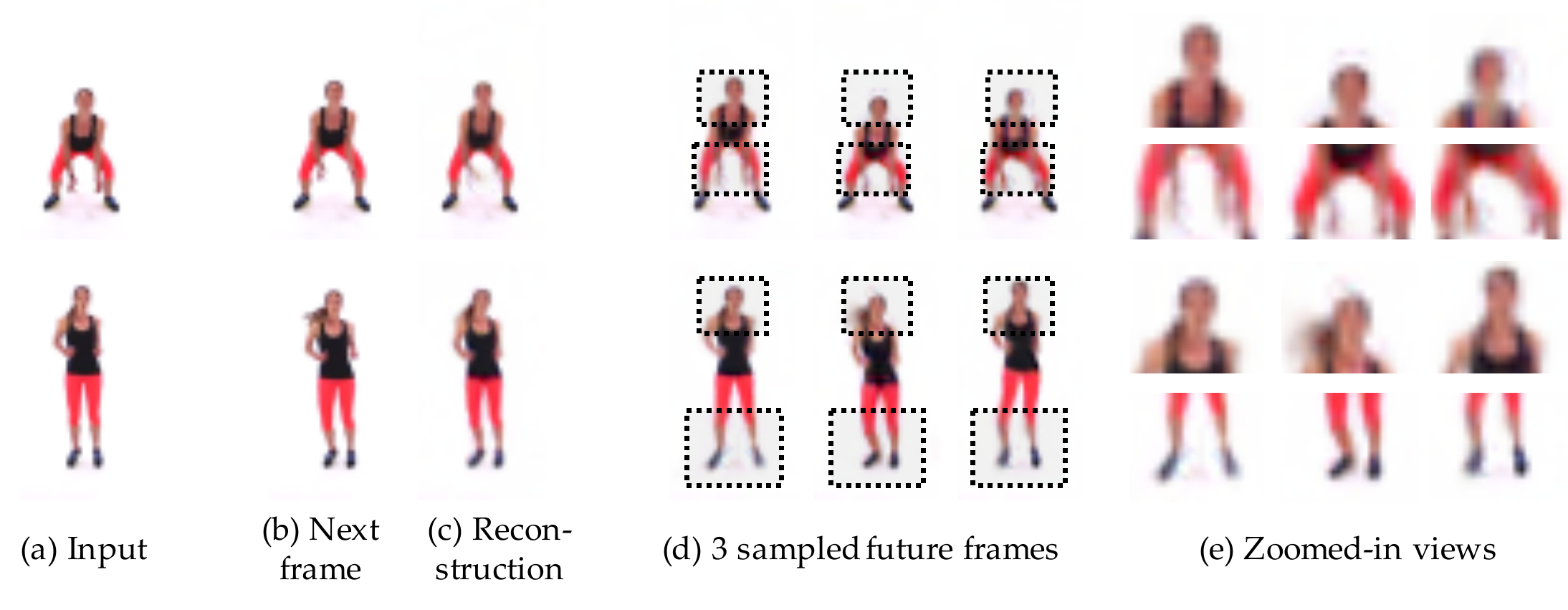} 
    \end{minipage}
    \hfill
    \begin{minipage}{0.27\linewidth}
        \centering
        \renewcommand{\arraystretch}{1.3} 
        \begin{tabular}{ccc}
            \toprule
            \multirow{2}{*}{Method} & \multicolumn{2}{c}{Resolution} \\
            \cmidrule{2-3} & 32$\times$32 & 64$\times$64 \\
            \midrule
            Flow & 31.3 & 25.5 \\
            Ours & \textbf{36.7} & \textbf{31.3} \\
            \bottomrule
        \end{tabular}
        
        \ \\
        \justifying\noindent
        \textit{\small Percentages (\%) of results labeled as \emph{real} by human subjects}
    \end{minipage}
    
    \caption{Left: Results on \textit{Exercise} dataset, where we show input images (a), ground truth next frames (b), our reconstruction (c), three sampled future frames (d), and corresponding zoomed-in views (e). Right: Percentages (\%) of synthesized results that were labeled as real by human subjects in two-way forced choices on Amazon Mechanical Turk, at resolution 32$\times$32 and 64$\times$64. A perfect algorithm would achieve a percentage around 50\%.}
    \label{fig:result_cardio}
\end{figure*}

%% file: figText/mask_kernel.tex
\begin{figure*}[t]
    \centering
    \includegraphics[width=0.49\linewidth]{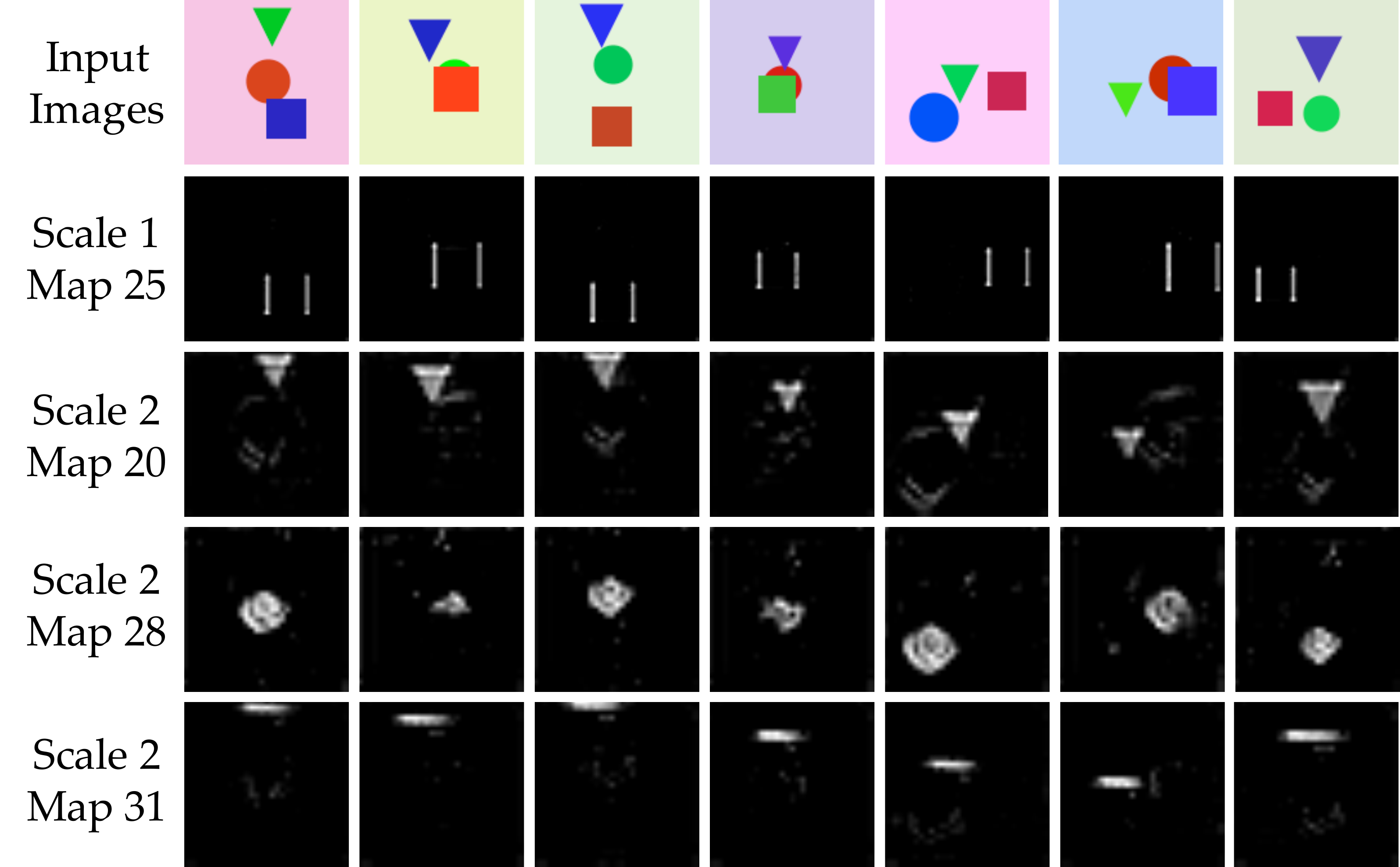}
    \hfill
    \includegraphics[width=0.49\linewidth]{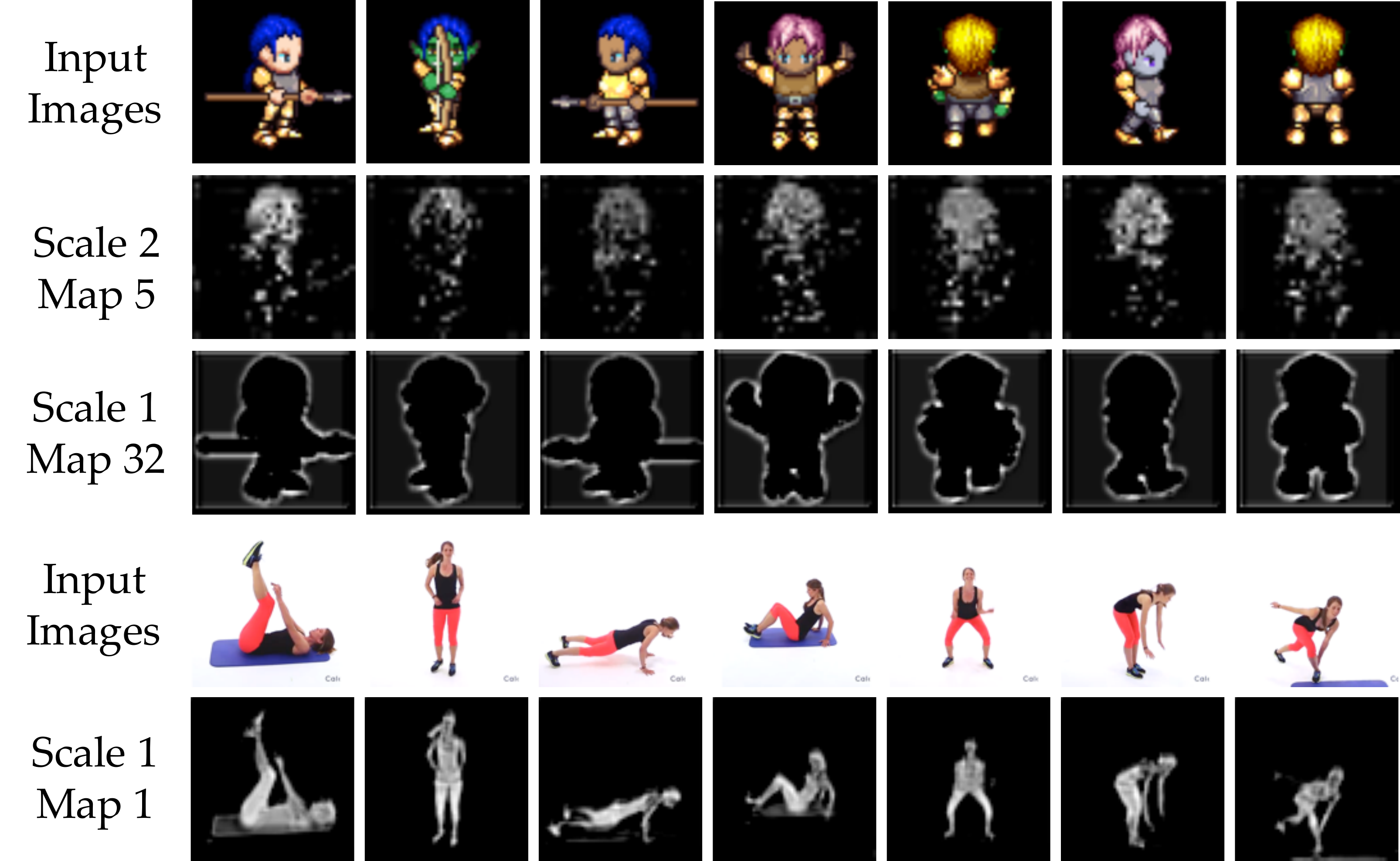}
    \caption{Learned layers on the Shapes dataset (left), the Sprites dataset (top right), and the Exercise dataset (bottom right). Our system is able to implicitly discover semantic structure from this self-supervised task. On the Shapes dataset, it learns to detect circles and triangles; it also detects vertical boundaries of squares, as squares always move horizontally. On the Exercise dataset, it learns to respond to only humans, not the carpet below them, as the carpet never moves. }
    \label{fig:mask}
\end{figure*}

%% file: figText/action.tex
\begin{figure}[t]
    \centering
    \includegraphics[width=\linewidth]{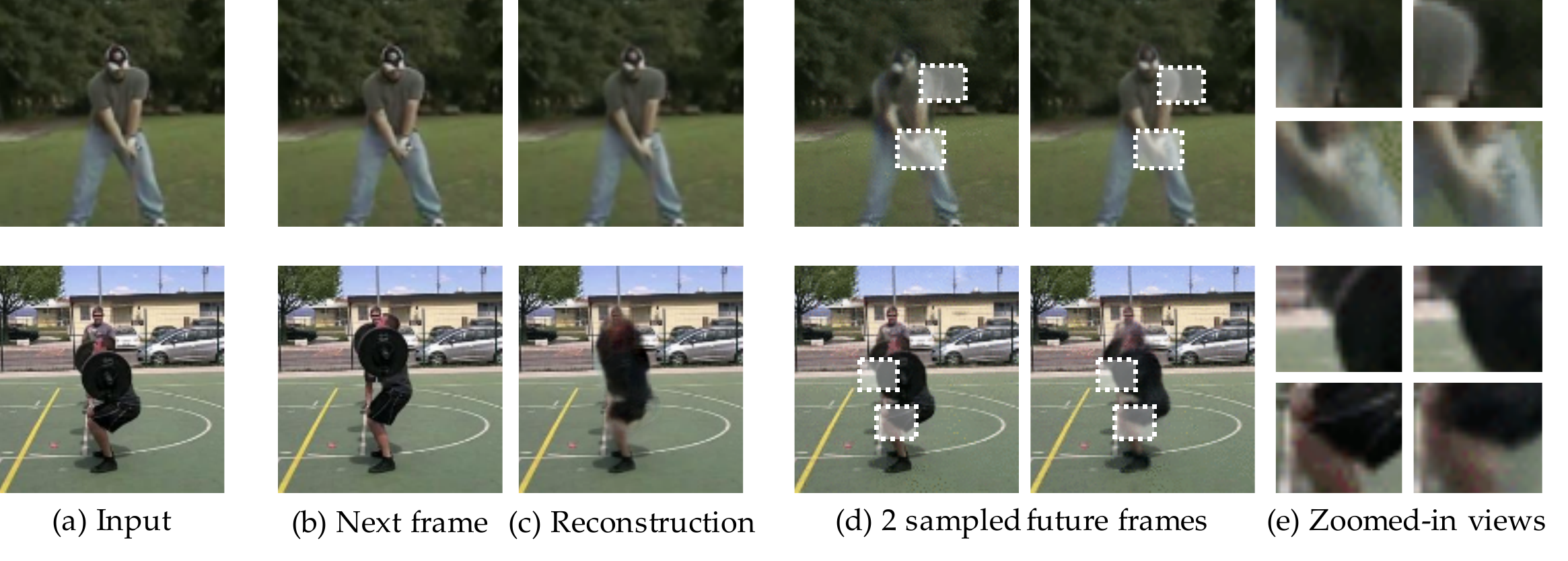} 
    \caption{Sampling results on the PennAction dataset~\cite{zhang2013actemes}, where we show input images (a), ground truth next frames (b), our reconstruction (c), three sampled future frames (d), and corresponding zoomed-in views (e).}
    \label{fig:result_action}
\end{figure}

%% file: figText/zstats_fig.tex
\begin{figure*}[t]
    \centering
    \includegraphics[width=\linewidth]{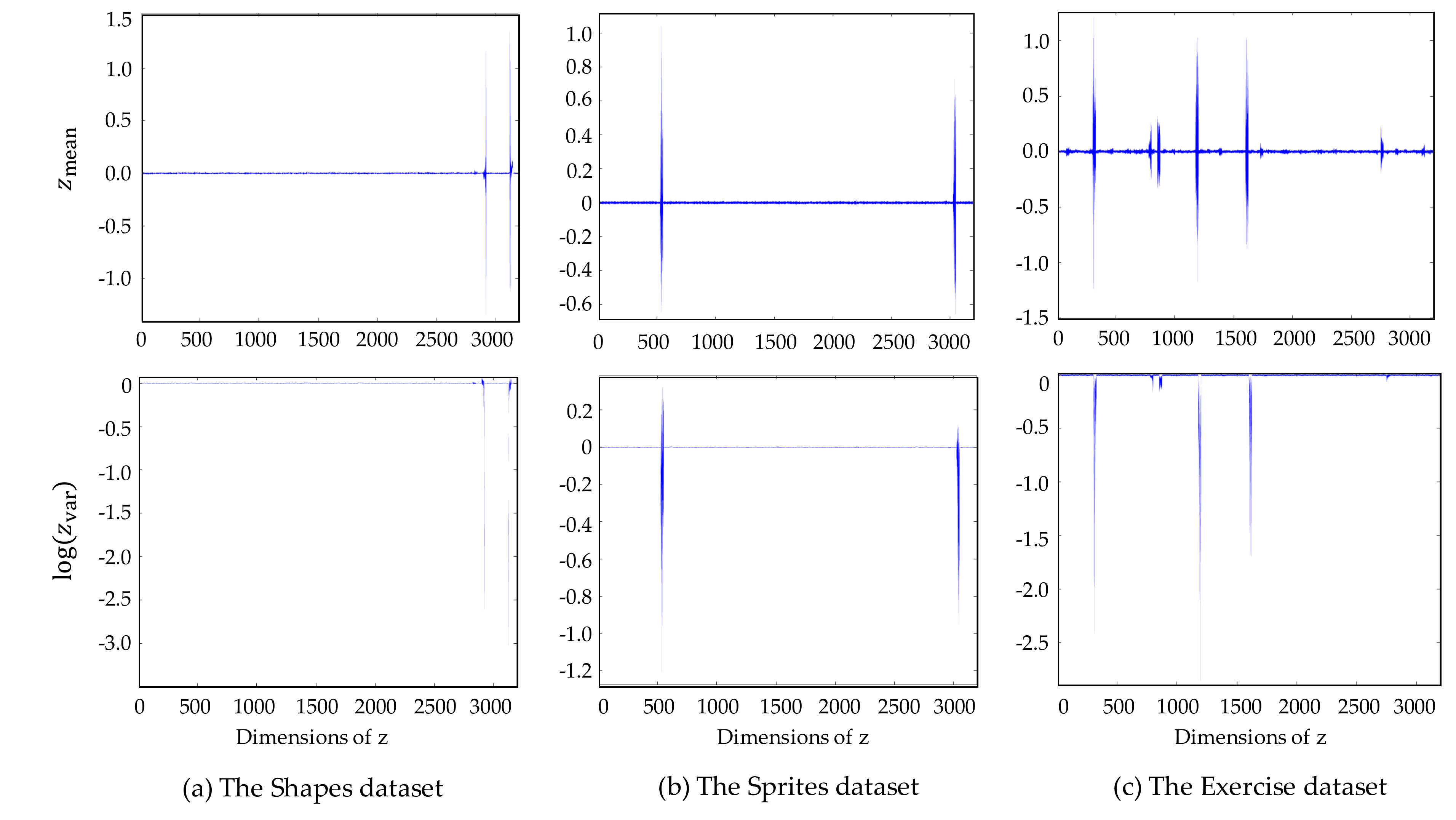}
    \caption{Statistics of latent vectors $z_\text{mean}$ and $z_\text{logvar}$ extracted from 1,000 image pairs from the Shapes, Sprites, and Exercise datasets, respectively. Each vertical line is for a single dimension in $z$. Although the $z$ vector has 3,200 dimensions, only a small number of those have values deviating from the prior (mean $0$, log-variance $0$). This shows the system is learning to encode motion information in a compact way. Also, as the motion in the Exercise dataset is more complex, the system needs more bits for encoding (\cf, \tbl{tbl:zstats}). } 
    \label{fig:zstats}
\end{figure*}

%% file: figText/zstats_tbl.tex
\begin{table}
    \centering
    \renewcommand{\arraystretch}{1.3} 
    \setlength{\tabcolsep}{4pt}
    \caption{Statistics of the 3,200-dimensional motion vector $z$}
    \label{tbl:zstats}
    
    \begin{tabular}{cccc}
    \toprule
        Dataset & Shapes & Sprites & Exercise \\
    \midrule
        Non-zero element in $z_{\text{mean}}$ & 299 & 54 & 978 \\
        Dominating PCA components & 5 & 5 & 47 \\
    \bottomrule
    \end{tabular}
    
    \ \\    
    \justifying\noindent
    \textit{\small The network learns a sparse latent representation, encoding high-level knowledge using minimal bits.}
\end{table}

%% file: text/analysis.tex
\section{Analyses}
\label{sec:analysis}

In this section, we present an analysis of the learned network to demonstrate what it captures. In particular, we visualize the learned feature maps (\sect{sec:vis_feature}), and show that the network implicitly learns object and edge detectors. Additionally, we compute statistics of the latent representation (\sect{sec:stats_z}) to verify that the network is learning a compact, informative representation of the motion manifold. We then visualize certain dimensions of the latent representation to reveal their meanings (\sect{sec:vis_z}). 

\subsection{Visualizing Learned Layers}
\label{sec:vis_feature}

Our network synthesizes the movement of objects by first identifying layers in the input and transferring each layer. Therefore, we expect that the these layers carry both low-level information such as object contours, and high-level information such as object parts.

To verify this, we visualize the learned feature maps (\fig{fig:pipeline}b) in \fig{fig:mask}. Even without supervision, our network learns to detect objects or contours in the image. For example, we see that the network automatically learns object detectors and edge detectors on the shapes dataset. It also learns a hair detector and a body detector on the sprites and exercise datasets, respectively. The part detectors have a sharper boundary on the shapes dataset than on the other two. This is because the shapes dataset has well-defined concepts of independently movable parts; in contrast, body parts in real videos have indistinguishable motion distributions.

\subsection{The Sparsity of the Latent Representation \texorpdfstring{$z$}{z}}
\label{sec:stats_z}

Although our latent motion representation, $z$, has 3,200 dimensions, its intrinsic dimensionality is much smaller. We show statistics in \tbl{tbl:zstats}. There are two main messages. First, $z_\text{mean}$ is very sparse. There are 299 non-zero elements of $z_\text{mean}$ in the shapes dataset, 54 in sprites, and 978 in exercise. The sprites dataset requires fewer non-zero elements than the shapes dataset, because while the visual appearance of the characters are more complex, their motion falls into a few pre-defined categories (\eg, thrust) and is thus simpler than the continuous motion of the shapes. Second, the independent components of $z$ are even fewer. We run principle component analysis (PCA) on the $z_\text{mean}$s obtained from a set of training images, and find that for each dataset, only a small fraction of components cover 95\% of the variance in $z_\text{mean}$ (5 in shapes, 5 in sprites, and 47 in exercise).

\fig{fig:zstats} further shows detailed statistics of $z_\text{mean}$ and $z_\text{logvar}$ (\aka, $\log(z_\text{var})$). For each dataset, we randomly select 1,000 train samples and calculate the corresponding $z_\text{mean}$ and $z_\text{logvar}$ through the motion encoder (\fig{fig:pipeline}a). The first row shows the distribution of $z_\text{mean}$ and the second row shows the distribution of $z_\text{logvar}$. The $x$-axis in each figure corresponds to the 3,200 dimensions in $z_\text{mean}$ and $z_\text{logvar}$. For each dimension $k$, the blue region reflects the interval 
\begin{equation}
[\text{mean}(z_k) - \text{std}(z_k), \text{mean}(z_k) + \text{std}(z_k)],
\end{equation}
where $\text{mean}(z_k)$ is the mean of the 1,000 values in the $k$ dimension for the 1,000 samples, and $\text{std}(z_k)$ is the standard deviation.

\input{figText/lambda}
\input{figText/vis_z}

One interesting observation from \fig{fig:zstats} is that, for most dimensions, $z_\text{mean}$ is very close to 0 and $z_\text{var}$ is very close to 1 ($z_\text{logvar}$ equals to 0). This is because the training object~\eqn{eq:sample_approx} minimizes the KL divergence between $N(\vec{0}, \boldsymbol{I})$ and $N(z_\text{mean}, \text{diag}(z_\text{var}))$\footnote{$\text{diag}(z_\text{var})$ denotes a diagonal matrix whose diagonal elements are $z_\text{var}$.}, and the minimizer of the KL divergence is $z_\text{mean} = \vec{0}$ and $z_\text{var} = \vec{1}$.

However, not all dimensions of $z_\text{mean}$ and $z_\text{logvar}$ are 0, and those non-zero dimensions actually encode the semantic information of how objects move between frames. Recall that to calculate the compact motion representation $z$, the motion encoder first estimates $z_\text{mean}$ and $z_\text{var}$ from two input frames, and then samples $z$ from $N(z_\text{mean}, \text{diag}(z_\text{var}))$. Let us rewrite the sampling as $z_k = z_{\text{mean},k} + \epsilon_k z_{\text{var},k}$, where $k$ is a dimension index and $\epsilon_k$ is a random Gaussian noise with zero mean and unit variance. If $z_{\text{var},k}$ is large, the corresponding dimension in the motion vector $z$ is mostly corrupted by the random Gaussian noise, and only those dimensions with small $z_{\text{var},k}$ are able to transmit motion information for next frame synthesis.

In other words, $z_\text{mean}$ carries the actual motion information, while $z_\text{var}$ is an indicator of whether a dimension is being actively used. This corresponds to our observation in \fig{fig:zstats}, where only uncorrupted dimensions (\ie, those with small $z_\text{var}$) have non-zero $z_\text{mean}$. Similar discussions were also presented by Hinton and van Camp~\cite{hinton1993keeping} and concurrently by Higgins~\etal\cite{higgins2016early}.

To further demonstrate how KL-divergence criterion ensures the sparsity of motion vector $z$, we also vary the weight of KL-divergence criterion $\lambda$ during training. As shown in \fig{fig:lambda}, when $\lambda$ is small, most of dimensions of $\log(z_\text{var})$ are smaller than 0, and $z_\text{mean}$ is not sparse (left of \fig{fig:lambda}). When we increase $\lambda$, the network is forced to encode the information in a more compact way, and the learned representation becomes sparser (right of \fig{fig:lambda}).

All these results suggest that our network has learned a compact representation of motion in an unsupervised fashion, and encodes high-level knowledge using a small number of bits, rather than simply memorizing training samples. In the next subsection, we will also illustrate what motion information is actually learned.

\input{figText/analogy}

\subsection{Varying the Size of the Latent Representation}
\label{sec:size_z}

We have explored the sparsity of the latent representation $z$, suggesting the model learns to encode essential information using only a few bits, despite $z$ itself has 3,200 dimensions. Here, we conduct an additional ablation study to understand the minimal number of dimensions that are necessary for the network to initialize the learning process.

We experiment on the \emph{Shapes} dataset. \tbl{tbl:zstats} tells us that our model can learn to encode motion information using only 5 dominating PCA components. We therefore explore 4 different sizes of $z$: 8, 32, 128, and 3,200, as shown in \fig{fig:size}. All networks converge regardless of the dimensions of $z$, with a slight difference in convergence time (the network with smaller z takes longer to converge). Also, the number of dominating PCA components is always small (3--5), suggesting that as long as the dimension of the latent representation $z$ is larger than its intrinsic dimension, the model consistently learns to encode the information compactly, regardless of its dimension.

\subsection{Visualizing the Latent Representation} 
\label{sec:vis_z}

We visualize the encoded motion by manipulating individual dimensions of the representation $z$, while keeping other dimensions constant. Through this, we have found that each dimension corresponds to a certain type of motion. We show results in \fig{fig:vis_z}. On the exercise dataset, varying one dimension of $z$ causes the girl to stand-up, and varying another causes her to move her leg. The effect is consistent across input images, showing individual dimensions in the latent vector $z$ carries abstract, higher-level knowledge. Also notice that only dimensions with smaller variance $z_\text{var}$ contains semantic motion information (\fig{fig:vis_z}a--c). Manipulating dimensions with variances close to 1 results in no significant motion (\fig{fig:vis_z}d). 

\subsection{Handling Disocclusions}
\label{sec:disocclude}

Synthesizing future frames often involves handling disocclusions. Here we systematically evaluate it on a new dataset, \emph{Shapes+Texture}, where the primitives in the \emph{Shapes} dataset now have horizontal (squares), vertical (triangles), or checkerboard patterns (circles). \fig{fig:disocclude} shows our model handles disocclusions well, correctly synthesizing the object texture even if it's not visible in the input. This is further supported by \fig{fig:mask}, showing the network learns feature maps that corresponding to \emph{amodal} segments of shapes.

\input{figText/tbl_analogy}

\input{figText/extrapolation}

%% file: figText/lambda.tex
\begin{figure*}[t!]
    \centering
    \begin{minipage}{.64\linewidth}
    \includegraphics[width=\linewidth]{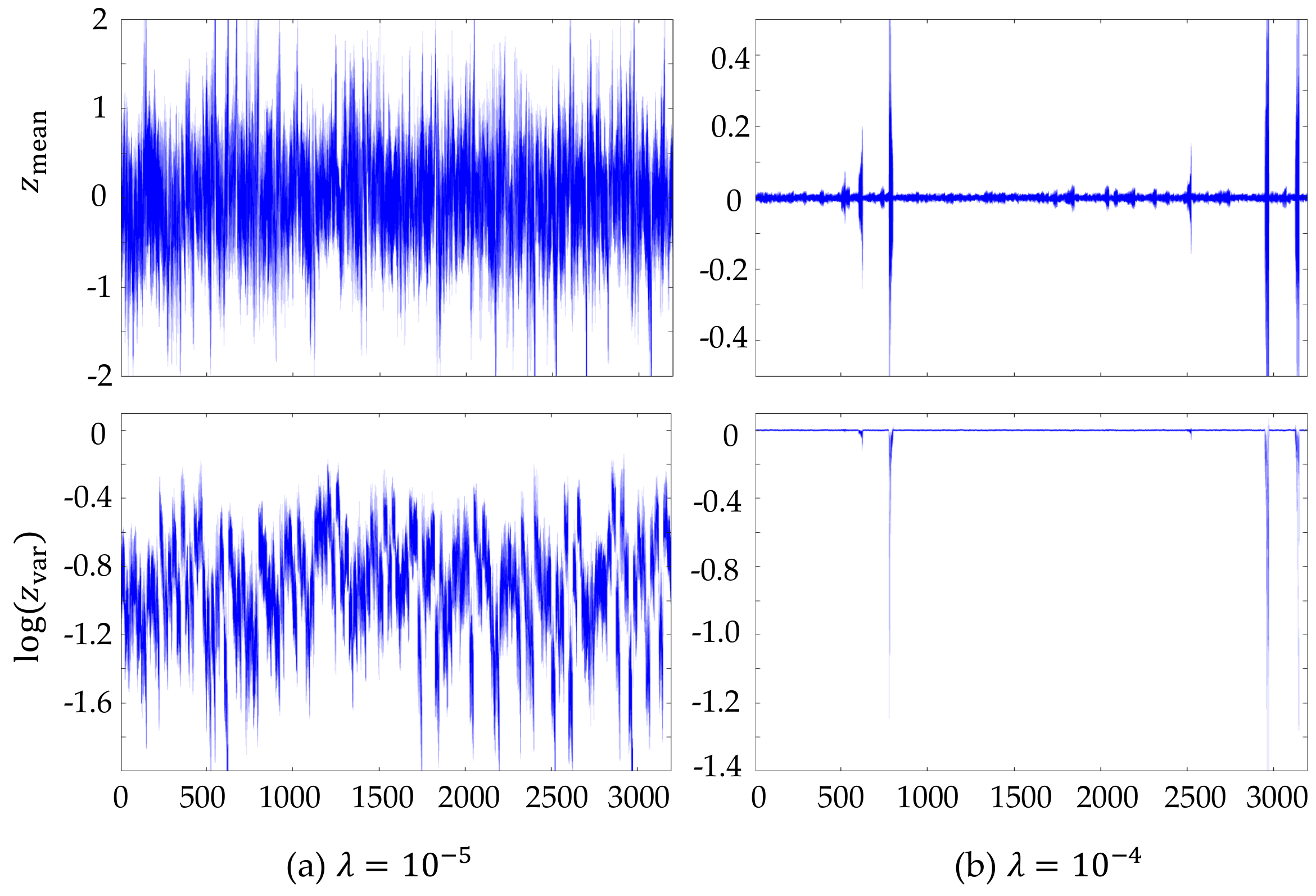}
    \vspace{-15pt} 
    \caption{Statistics of latent vectors $z$ extracted from 1,000 image pairs from the Shapes dataset with networks learned using different values of $\lambda$. The latent vector becomes sparser when $\lambda$ is larger, \ie, the network is encoding motion in a more compact way. 
    }
    \label{fig:lambda}
    \end{minipage}
    \hfill
    \begin{minipage}{.33\linewidth}
    \centering
    \includegraphics[width=\linewidth]{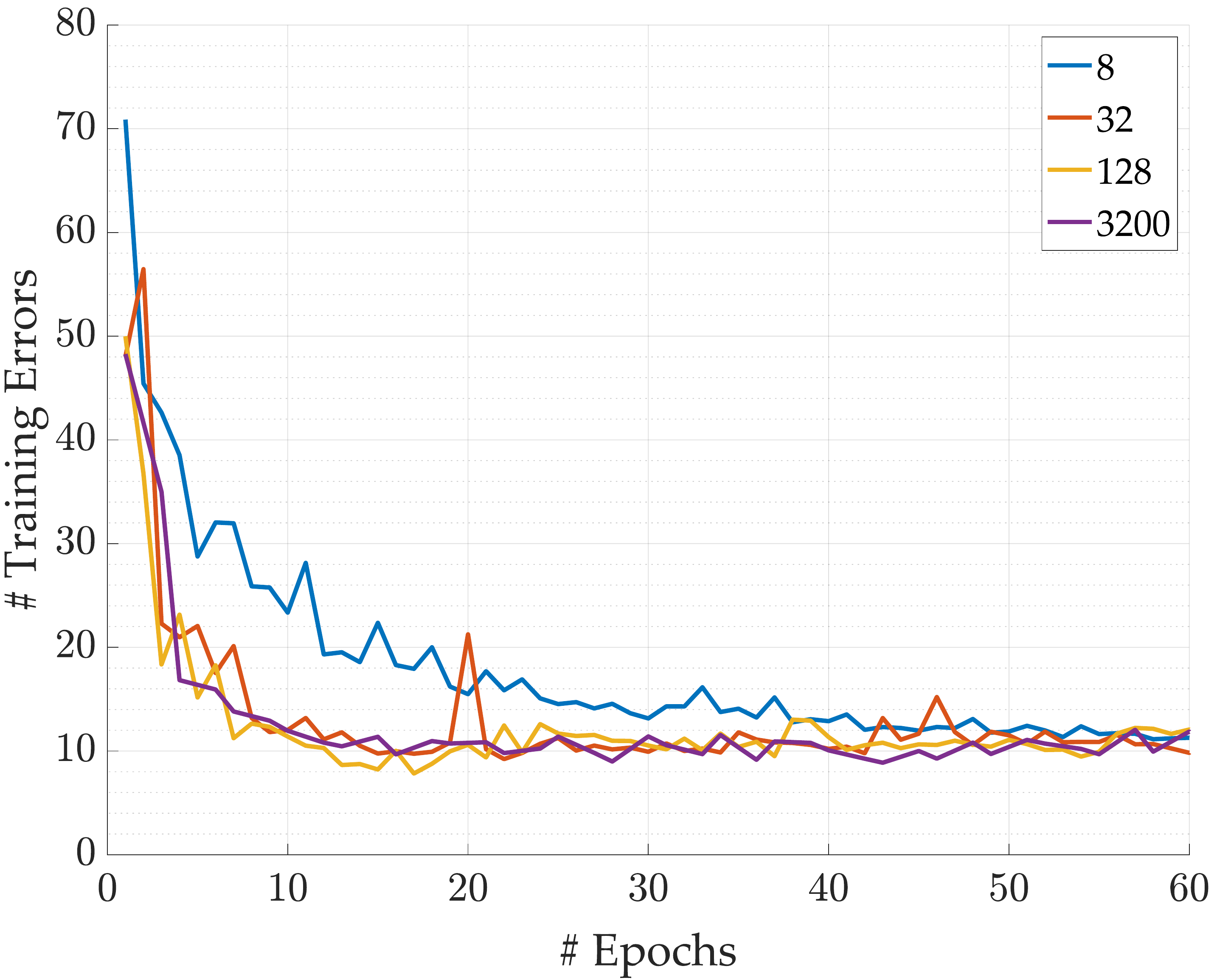}
    
    \ \\
    \renewcommand{\arraystretch}{1.3} 
    \setlength{\tabcolsep}{4pt}
    \begin{tabular}{lcccc}
    \toprule
        Size of $z$ & 8 & 32 & 128 & 3,200 \\
    \midrule
        \specialcell{Dominating\\PCA components} & 3 & 3 & 3 & 5 \\
    \bottomrule
    \end{tabular}
    \vspace{-5pt} 
    \caption{Ablation study on the size of $z$. Our model automatically discovers the underlying dimension of the motion in the \emph{Shapes} dataset. When we shrink the size of $z$, the convergence gets a little slower (especially when $z$ has only 8 dimensions), but the results are essentially the same.}
    \label{fig:size}
    \end{minipage}
\end{figure*}

%% file: figText/vis_z.tex
\begin{figure*}[t!]
    \centering
    \includegraphics[width=\linewidth]{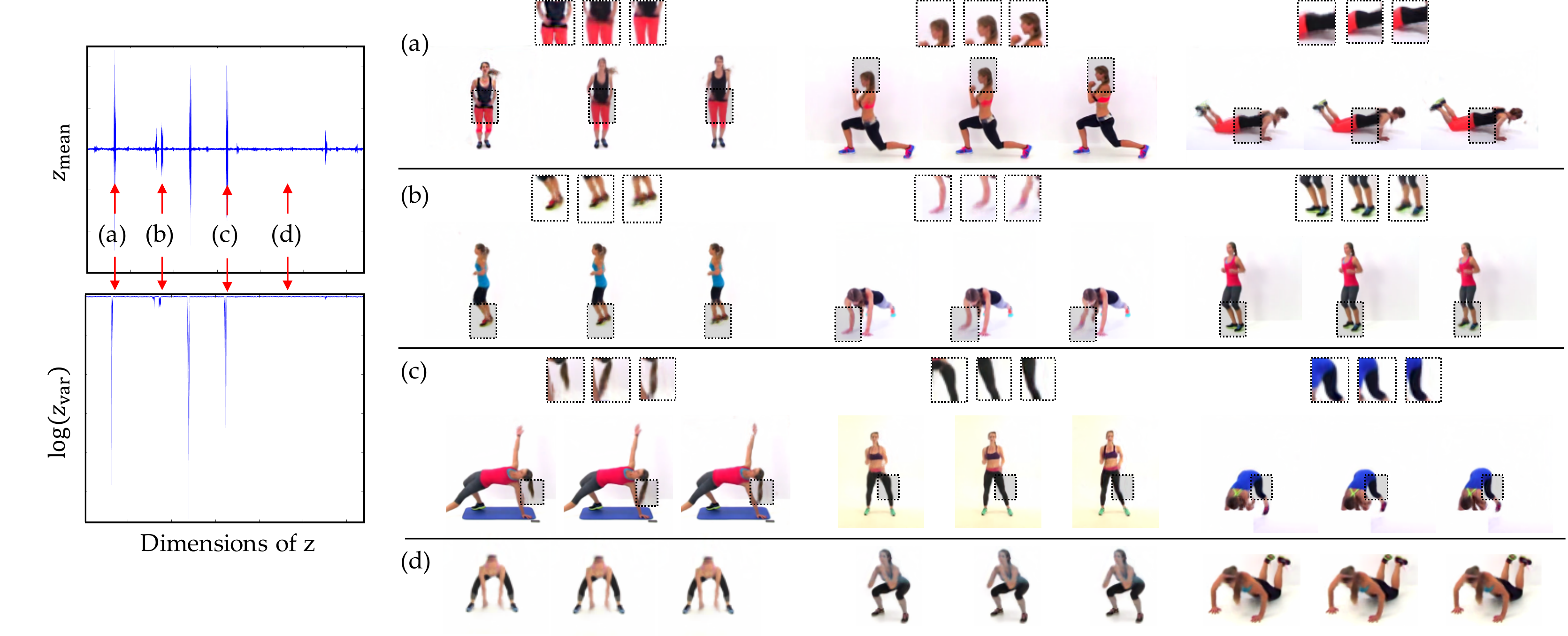}
    \vspace{-10pt}
    \caption{We visualize the effect of varying individual dimensions in the latent representation $z$, revealing the system is learning a disentangled, interpretable motion representation. For the dimensions whose log-variance is smaller than $0$ (a--c), they record a certain type of motion. For example, dimension (a) corresponds to humans move upwards, dimensions (b) and (c) correspond to moving arms, hair, or legs to the left. For the dimensions whose log-variance is very close to $0$, they record no motion information: changing the value of dimension (d) results in no motion in predicted frames.}
    
    \label{fig:vis_z}
\end{figure*}

%% file: figText/analogy.tex
\begin{figure*}[t]
\centering
\begin{minipage}{.32\linewidth}
    \includegraphics[width=\linewidth]{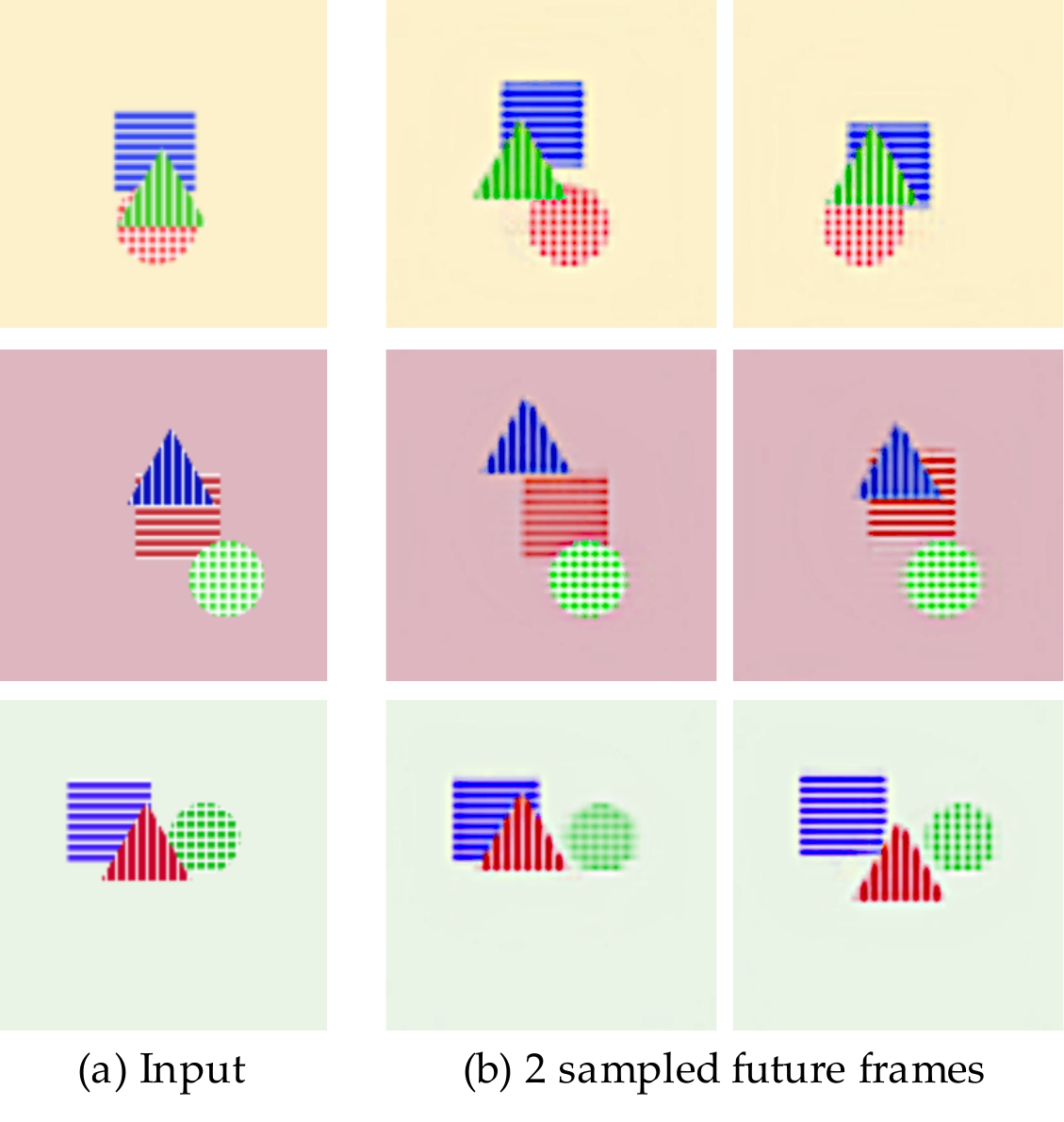}
    \vspace{-15pt}
    \caption{Our model handles disocclusions well. On the \emph{Shapes+Texture} dataset, our model is able to complete shapes with their corresponding texture after hallucinating their motion.}
    \label{fig:disocclude}
\end{minipage}
\hfill
\begin{minipage}{.66\linewidth}
    \setlength{\fboxrule}{1.5pt}
    \centering
    \begin{tabular}{C{0.14\linewidth}C{0.80\linewidth}}
    Reference &
    \includegraphics[width=0.15\linewidth]{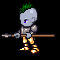}
    \includegraphics[width=0.15\linewidth]{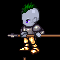}
    $\;$
    \includegraphics[width=0.15\linewidth]{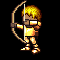}
    \includegraphics[width=0.15\linewidth]{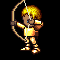}
    $\;$
    \includegraphics[width=0.15\linewidth]{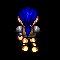}
    \includegraphics[width=0.15\linewidth]{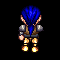}
    \\
    Input + {\color{red} Prediction} & 
    \includegraphics[width=0.15\linewidth]{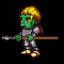}
    \includegraphics[width=0.15\linewidth,cframe=red {\fboxrule} {-\fboxrule}]{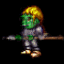}
    $\;$
    \includegraphics[width=0.15\linewidth]{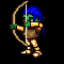}
    \includegraphics[width=0.15\linewidth,cframe=red {\fboxrule} {-\fboxrule}]{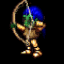}
    $\;$
    \includegraphics[width=0.15\linewidth]{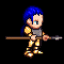}
    \includegraphics[width=0.15\linewidth,cframe=red {\fboxrule} {-\fboxrule}]{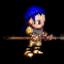}
    \\
    Reference &
    \includegraphics[width=0.15\linewidth]{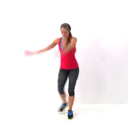}
    \includegraphics[width=0.15\linewidth]{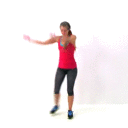}
    $\;$
    \includegraphics[width=0.15\linewidth]{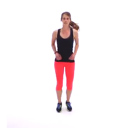}
    \includegraphics[width=0.15\linewidth]{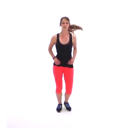}
    $\;$
    \includegraphics[width=0.15\linewidth]{fig/analogy/6A-1.png}
    \includegraphics[width=0.15\linewidth]{fig/analogy/6A-2.png}
    \\
    Input + {\color{red} Prediction} & 
    \includegraphics[width=0.15\linewidth]{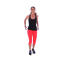}
    \includegraphics[width=0.15\linewidth,cframe=red {\fboxrule} {-\fboxrule}]{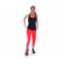}
    $\;$
    \includegraphics[width=0.15\linewidth]{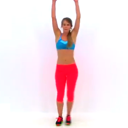}
    \includegraphics[width=0.15\linewidth,cframe=red {\fboxrule} {-\fboxrule}]{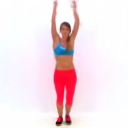}
    $\;$
    \includegraphics[width=0.15\linewidth]{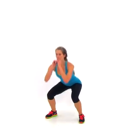}
    \includegraphics[width=0.15\linewidth,cframe=red {\fboxrule} {-\fboxrule}]{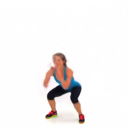}
    \end{tabular}
    \caption{Results on visual analogy-making, where we want to transfer the motion in a reference pair to a target image. We conduct experiments on two datasets, Sprites and Exercise, and mark the predicted frames in red. Our algorithm is able to transfer high-level motion (\eg downwards motion of a human body, as opposed to pixel-wise motion) in a semantically plausible way.}
    \label{fig:analogy}
\end{minipage}
    \vspace{-5pt}
\end{figure*}

%% file: figText/tbl_analogy.tex
\begin{table}[t]

    \centering
    \renewcommand{\arraystretch}{1.3} 
    \setlength{\tabcolsep}{4pt}
    \caption{Mean squared pixel error on test analogies, by animation}
    \vspace{-10pt}
    \label{tbl:analogy}
    
    \begin{tabular}{lcccccc}
        \toprule
             Model & spellcast & thrust & walk & slash & shoot & average \\
        \midrule
            Add & 41.0 & 53.8 & 55.7 & 52.1 & 77.6 & 56.0 \\
            Dis & 40.8 & 55.8 & 52.6 & 53.5 & 79.8 & 56.5 \\
            Dis+Cls & 13.3 & 24.6 & 17.2 & {\bf 18.9} & 40.8 & 23.0 \\
        \midrule
            Ours & {\bf 9.5} & {\bf 11.5} & {\bf 11.1} & 28.2 & {\bf 19.0} & {\bf 15.9} \\
        \bottomrule
    \end{tabular}
    
    \ \\
    \justifying\noindent
    \textit{\small The first three models (Add, Dis, and Dis+Cls) are from Reed~\etal~\cite{reed2015deep}. Compared with them, our model achieves lower errors.}
\end{table}

%% file: figText/extrapolation.tex
\begin{figure*}[t]
    \centering
    \includegraphics[width=\linewidth]{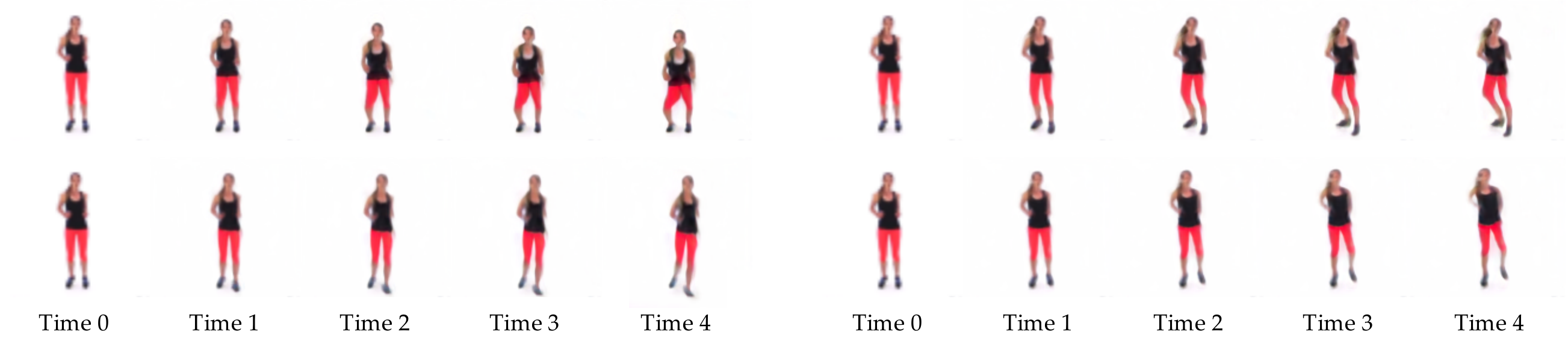}
    \vspace{-5pt}
    \caption{Generating video sequences by repeatedly applying a sampled motion representation $z$. Our model is able to synthesize reasonable videos of human moving in various ways, though artifacts gradually emerge over time (\eg the thighs become bigger in the top-right example). 
    }
    \vspace{-5pt}
    \label{fig:extrapolation}
\end{figure*}

%% file: text/application.tex
\section{Applications}
\label{sec:app}

Our framework models a general problem and therefore has a wide range of applications. Here we present two possible applications of our framework: visual analogy-making and extrapolation for generating video sequences.

\subsection{Zero-Shot Visual Analogy-Making}
\label{sec:analogy}

Recently, Reed~\etal~\cite{reed2015deep} studied the problem of inferring the relationship between a pair of reference images and synthesizing an image analogy by applying the inferred relationship to a test image. For example, the character shown in the top row of \fig{fig:analogy}a leans toward to the right; the task is to transfer its leaning motion to the target, bottom-left image.

The method by Reed~\etal requires a set of quadruples during supervision (two source images and two target images). In contrast, our network is able to preform this task without first training using the quadruple sets. Specifically, we extract the motion vector, $z$, from two reference frames using our motion encoder (\fig{fig:pipeline}a). We then use the extracted motion vector $z$ to synthesize an analogy-image given a new test image. In this way, our network learns to transfer motion from a source pair to the target image, without requiring any form of supervision (see \fig{fig:analogy}). As shown in \tbl{tbl:analogy}, our algorithm out-performs that by Reed~\etal~\cite{reed2015deep}.

\subsection{Extrapolation}
\label{sec:extrapolation}

Our network can also be used to generate video sequences by extrapolation. In \fig{fig:extrapolation}, we synthesize videos from a single image by simply repeatedly applying the sample motion. Given an input image frame $I_1$ and sampled motion vector $z$, we first synthesize the second frame $I_2$ from the first frame $I_1$ and motion vector $z$, and then synthesize the third frame $I_3$ from the second using the same motion vector. 
We see that our framework generates plausible video sequences. Recent work on modeling transitions across possible motions~\cite{chao2017forecasting}, could serve as alternative way to extend our framework to longer-term video generation. Compared with most of previous deterministic video synthesis networks~\cite{srivastava2015unsupervised,mathieu2015deep}, our approach can sample multiple possible ways a person can move, as shown \fig{fig:extrapolation}. One limitation of this approach is that artifacts gradually emerge over time. It may be possible to reduce these artifacts using a learned image prior~\cite{zhang2017real}.

%% file: text/conclusion.tex
\section{Conclusion}
\label{sec:conclusion}

In this paper, we have proposed a novel framework that samples future frames from a single input image. Our method incorporates a variational autoencoder for learning compact motion representations and layer-based synthesis algorithm to generate realistic movement of objects. We have demonstrated that our framework works well on both synthetic and real-life videos. 

The key component of our frame synthesis model is to decompose an input image into different layers and model the movement of each layer through a simple convolutional network. This motion can well model the motion of a single deformable object, like human body. In the future, we would also like to extend it to handle more complicated motion and stochastic motion, \eg, water flowing.

Current results also suggest that our probabilistic visual dynamics model may be useful for additional applications, such as inferring objects' higher-order relationships by examining correlations in their motion distributions. Furthermore, this learned representation could be potentially used as a sophisticated motion prior in other computer vision and computational photography applications.

%% file: text/appendix.tex
\section{Verification of \eqn{eq:sample_approx} and \eqn{eq:final_bound}}
\label{sec:appendix}

In this section, we formally derive how we obtain the training objective function in \eqn{eq:sample_approx}, following similar derivations in \cite{kingma2013auto,kingma2014semi,yan2015attribute2image}. As mentioned in \sect{sec:vae}, the generative process that samples a difference image $v$ from a $\theta$-parametrized model, conditioned on an observed image $I$, consists of two steps. First, the algorithm samples the hidden variable $z$ from a prior distribution $p_z(z)$. Then, given a value of $z$, the algorithm samples the intensity difference image $v$ from the conditional distribution $p_\theta(v|I,z)$. This process is also described in the graphical model in \fig{fig:graphical_model}d.

Given a set of training pairs $\{I\i,v\i\}$, the algorithm maximizes the log-likelihood of the conditional marginal distribution during training
\begin{equation}
\sum_i \log p(v^{(i)}|I\i).\label{eqn:marginal1}
\end{equation}
Recall that $I$ and $z$ are independent as shown in the graphical model in \fig{fig:graphical_model}. Therefore, based on the Bayes' theorem, we have
\begin{equation}
\pv = \frac{\pz \ptheta}{\pzvi}.
\label{eqn:marginal2}
\end{equation}
It is hard to directly maximizing the marginal distribution in \eqn{eqn:marginal1}. We therefore maximize its variational upper-bound instead, as proposed by Kingma and Welling~\cite{kingma2013auto}. Let $q_\phi(z|v\i,I\i)$ be the variational distribution that approximates the posterior $p(z|v\i,I\i)$. Then each term in the marginal distribution is upper bounded as
\begin{align}
& \log \pv \notag \\
= & \mathbb{E}_{q_\phi}\left[\log \pv \right] \notag\\
= & \mathbb{E}_{q_\phi}\left[\log \frac{\pz \ptheta}{\pzvi}\right] \notag\\
= & \mathbb{E}_{q_\phi}\left[\log \frac{\pz}{\qphi}\right] + \mathbb{E}_{q_\phi}\left[\log \frac{\qphi}{\pzvi}\right] \notag \\
  \quad & + \mathbb{E}_{q_\phi}[\log \ptheta] \notag\\
= & -D_\text{KL}(\qphi || \pz) \notag \\
  \quad & + D_\text{KL}(\qphi || \pzvi) \notag \\
  \quad & + \mathbb{E}_{q_\phi}[\log \ptheta] \notag\\
\geq & -D_\text{KL}(\qphi || \pz) + \mathbb{E}_{q_\phi}[\log \ptheta] \notag\\
\overset{\Delta}{=} & \cL(\theta,\phi,v\i|I\i)
\label{eq:variational_bound}
\end{align}

The first KL-divergence term in \eqn{eq:variational_bound} has an analytic form~\cite{kingma2013auto}. To make the second term tractable, we approximate the \regdis,  $q_\phi(z|x\i,I\i)$, by its empirical distribution. We have
\begin{align}
& \cL(\theta,\phi,v\i|I\i) \notag \\
\approx & - D_{\text{KL}}(q_\phi(z|v\i,I\i) || p_z(z)) \notag \\
\quad & + \frac{1}{L} \sum_{l=1}^{L} \log p_\theta(v\i|z^{(i,l)},I\i) \notag \\
= & - D_{\text{KL}}(q_\phi(z|v\i,I\i) || p_z(z)) \notag \\
\quad & - \frac{1}{2 \sigma^2 L} \sum_{l=1}^{L} \|v\i - f_{\mbox{\tiny{mean}}}(z^{(i)},I\i)\| + C, \label{eq:sample_approx2}
\end{align}
where $z^{(i,l)}$ are samples from the \regdis $\qphi$ and $C$ is a constant. \eqn{eq:sample_approx2} is the variation lower bound that our network minimizes during training.

In practice, we simply generate one sample of $z^{(i,l)}$ at each iteration (thus $L=1$) of stochastic gradient descent, and different samples are used for different iterations. By defining $\lambda = 1/(2 \sigma^2)$ and taking the negative of the right-hand side of \eqn{eq:sample_approx2} , we get the objective function to minimize in training (\eqn{eq:final_bound}):
\begin{align}
\vspace{-10pt} D_{\text{KL}}(q_\phi(z|v\i,I\i) || p_z(z)) + \lambda \|v\i - f_{\mbox{\tiny{mean}}}(z^{(i)},I\i)\|.
\end{align}